\documentclass[10pt,twocolumn, letterpaper]{article}

\usepackage{cvpr} 
\usepackage{multirow}
\usepackage{bbding}
\usepackage{pifont}
\usepackage{xcolor}
\usepackage{colortbl}
\usepackage{algorithm}
\usepackage{algpseudocode}

\definecolor{cvprblue}{rgb}{0.21,0.49,0.74}
\usepackage[pagebackref,breaklinks,colorlinks,allcolors=cvprblue]{hyperref}
\def\paperID{xxxx} 
\def\confName{CVPR}
\def\confYear{2026}
\definecolor{jade}{RGB}{0,120,0}
\definecolor{newgreen}{RGB}{0, 153, 0}

\newcommand{\first}[1]{{\textcolor{red}{\textbf{#1}}}}     
\newcommand{\second}[1]{{\textcolor{blue}{\underline{#1}}}}

\title{AlignVAR: Towards Globally Consistent Visual Autoregression for Image Super-Resolution}

\author{
Cencen Liu$^1$, \and
Dongyang Zhang$^{1, 2}$, \and
Wen Yin$^1$, \and
Jielei Wang$^{1, 2}$, \and
Tianyu Li$^1$, \and
Ji Guo$^1$, \and
Wenbo Jiang$^1$, \and
Guoqing Wang$^1$, \and
Guoming Lu$^{1, 2, *}$ \and
$^1$ University of Electronic Science and Technology of China\\
$^2$ Ubiquitous Intelligence and Trusted Services Key Laboratory of Sichuan Province\\
}

\begin{document}
\maketitle

\begin{abstract}
Visual autoregressive (VAR) models have recently emerged as a promising alternative for image generation, offering stable training, non-iterative inference, and high-fidelity synthesis through next-scale prediction. This encourages the exploration of VAR for image super-resolution (ISR), yet its application remains underexplored and faces two critical challenges: \textbf{locality-biased attention}, which fragments spatial structures, and \textbf{residual-only supervision}, which accumulates errors across scales, severely compromises global consistency of reconstructed images. To address these issues, we propose \textbf{AlignVAR}, a globally consistent visual autoregressive framework tailored for ISR, featuring two key components: \emph{(1) Spatial Consistency Autoregression (SCA)}, which applies an adaptive mask to reweight attention toward structurally correlated regions, thereby mitigating excessive locality and enhancing long-range dependencies; and \emph{(2) Hierarchical Consistency Constraint (HCC)}, which augments residual learning with full reconstruction supervision at each scale, exposing accumulated deviations early and stabilizing the coarse-to-fine refinement process. Extensive experiments demonstrate that AlignVAR consistently enhances structural coherence and perceptual fidelity over existing generative methods, while delivering over 10× faster inference with nearly 50\% fewer parameters than leading diffusion-based approaches, establishing a new paradigm for efficient ISR.


\end{abstract}    
\section{Introduction}
\label{sec:intro}

\begin{figure}[!t]
    \centering
    \includegraphics[width=1\linewidth]{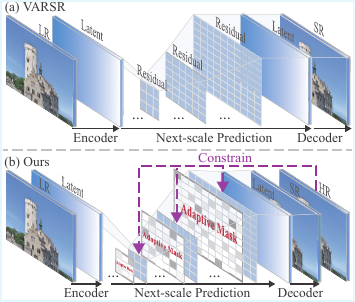}
    \caption{Comparison between the VARSR and AlignVAR. AlignVAR enhances VAR by introducing an adaptive consistency mask for intra-scale modeling and full reconstruction supervision for inter-scale alignment.}
    \vspace{-10pt}
    \label{fig:compare}
\end{figure}
Generative models, empowered by strong learned priors, have revolutionized the field of image super-resolution (ISR). Among them, Generative Adversarial Networks (GANs)~\cite{Ledig2016SRGAN, li2022best, ma2024uncertainty} and Diffusion Models~\cite{yang2024pasd, wang2024StableSR, lin2024diffbir} have emerged as the two dominant paradigms. However, each of these approaches suffers from inherent limitations. GAN-based methods~\cite{wang2018esrgan,Ledig2016SRGAN, zhang2019ranksrgan,yuan2018unsupervised}, while enhancing perceptual realism, often exhibit training instability and tend to introduce visually inconsistent artifacts~\cite{wu2024seesr}. Diffusion-based approaches~\cite{wu2024seesr,wang2024StableSR,lin2024diffbir,chen2025AdcSR,yang2024pasd} exploit powerful generative priors to achieve high-fidelity reconstruction, yet their iterative denoising process incurs heavy computational costs, substantially limiting their practicality~\cite{aithal2024hallucination}.

These challenges have motivated the exploration of alternative generative paradigms for ISR, among which visual autoregressive (VAR) modeling has recently emerged as a promising direction~\cite{tian2024var, wei2025pure}. The coarse-to-fine prediction strategy of VAR naturally aligns with the hierarchical nature of ISR, which progressively restores image details across scales. The recent VARSR~\cite{qu2025varsr}, as illustrated in Fig.~\ref{fig:compare}, pioneers the application of VAR to ISR, demonstrating its initial feasibility. However, this seminal attempt also reveals a fundamental conflict: the model’s inherent locality bias and the cumulative error propagation across scales jointly undermine the global consistency required for ISR, resulting in degraded global coherence in the reconstructed images.

\begin{figure}[!t]
    \centering
    \includegraphics[width=1\linewidth]{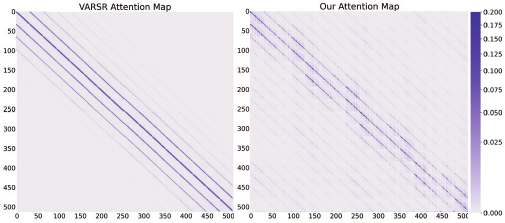}
    \caption{Comparison of attention distribution. Visualization of attention maps for VARSR and AlignVAR shows that VARSR exhibits highly localized attention concentrated in nearby regions, whereas AlignVAR captures broader contextual dependencies through the proposed Spatial Consistency Autoregression (SCA), thereby enhancing spatial coherence within each scale.}
    \vspace{-20pt}
    \label{fig:attention}
\end{figure}

\begin{figure}[!t]
    \centering
    \includegraphics[width=1\linewidth]{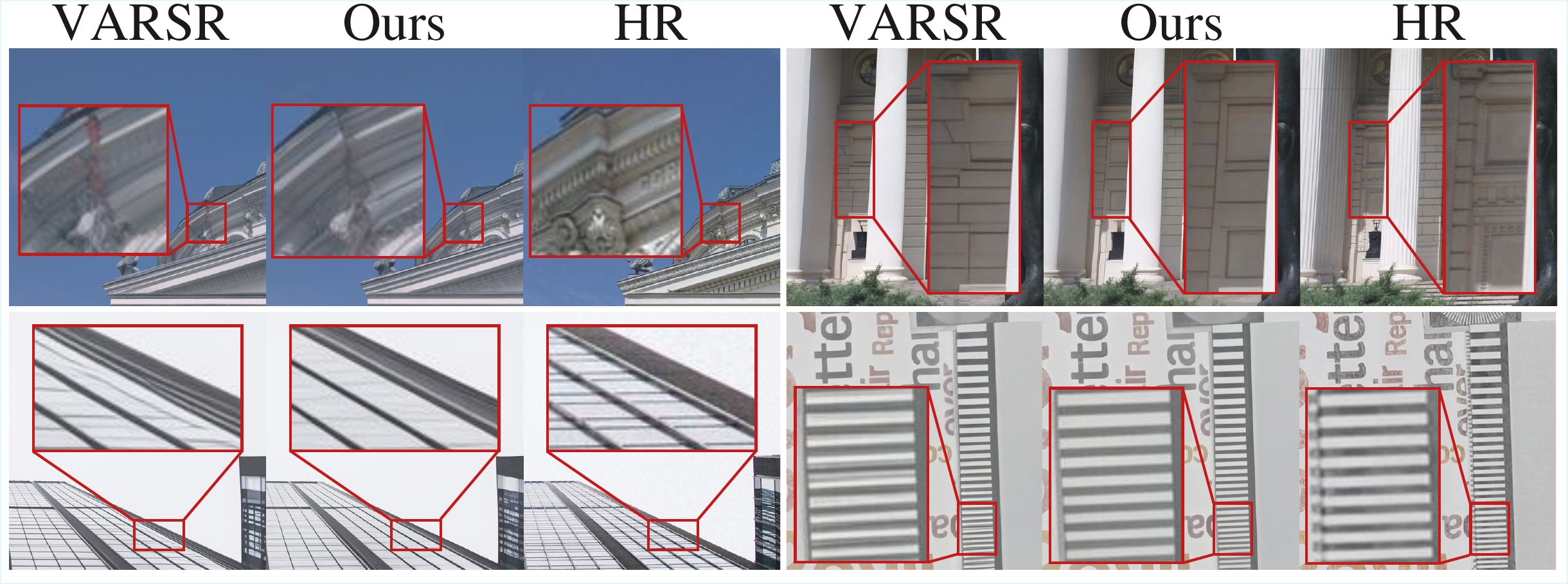}
    \vspace{-15pt}
    \caption{Spatial inconsistency results in texture discontinuities, structural distortions.}
    \vspace{-10pt}
    \label{fig:motivation-a}
\end{figure}

\begin{figure}[!t]
    \centering
    \includegraphics[width=1\linewidth]{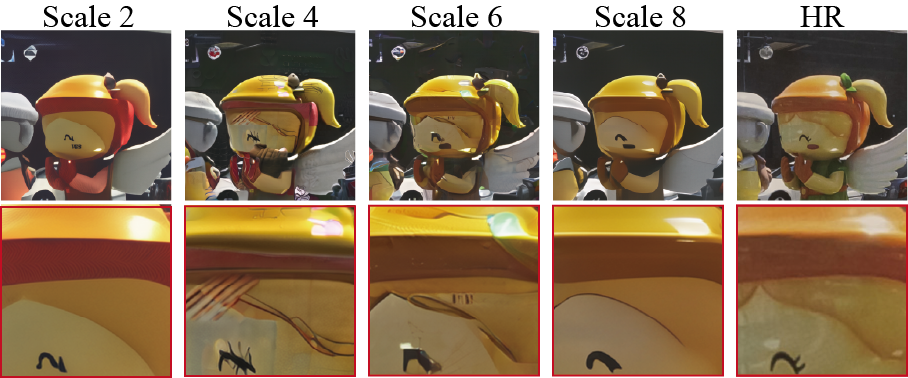}
    \vspace{-15pt}
    \caption{Hierarchical inconsistency results in color shifts and structural misalignment.}
    \vspace{-15pt}
    \label{fig:motivation-b}
\end{figure}

Building on this observation, we conduct a systematic analysis of the VARSR framework and identify two coupled root causes underlying its failure to maintain global consistency. The first is \textit{\textbf{Spatial Inconsistency}}. The self-attention mechanism in VAR models exhibits a strong locality bias, with attention weights concentrated almost exclusively on adjacent regions, as illustrated in Fig.~\ref{fig:attention}. This restricted receptive field limits the integration of global context, resulting in spatially disjoint artifacts such as fragmented textures and structural distortions, as shown in Fig.~\ref{fig:motivation-a}. The second is \textit{\textbf{Hierarchical Inconsistency}}, which we analyze by injecting random perturbations into different scales and observing their influence on the final reconstruction, as shown in Fig.~\ref{fig:motivation-b}. These perturbations induce color shifts and structural misalignments, indicating that residual-only supervision allows small prediction errors from coarser scales to propagate and amplify through the hierarchy~\cite{kumbong2025hmar}. We argue that these issues stem from a common origin: the lack of explicit consistency constraints in both the intra-scale (spatial) and inter-scale (hierarchical) dimensions.

To address these limitations, we propose \textbf{AlignVAR}, a visual autoregressive framework designed to achieve globally consistent ISR. As shown in Fig.~\ref{fig:compare}, AlignVAR introduces two complementary mechanisms that collaboratively promote coherence within and across scales by reweighting spatial attention and recalibrating hierarchical dependencies. Specifically, the \textit{Spatial Consistency Autoregression (SCA)} adaptively reweights attention to emphasize structurally correlated regions rather than local neighborhoods, enabling the model to aggregate long-range context and maintain spatial continuity. Meanwhile, the \textit{Hierarchical Consistency Constraint (HCC)} recalibrates cross-scale dependencies through a hierarchical supervision objective that enforces both residual and full-scale consistency between predictions and ground-truth representations. By allowing each scale to correct contextual deviations before they propagate, HCC effectively suppresses error accumulation and stabilizes refinement throughout the hierarchy. SCA and HCC form a unified consistency-driven autoregressive formulation that restores fine textures while preserving large-scale structural fidelity, yielding reconstructions that are both perceptually accurate and globally coherent.


In summary, we make the following key contributions:
\begin{itemize}
    \item We revisit visual autoregression for ISR and identify two fundamental sources of inconsistency in existing VAR-based frameworks: \textit{spatial inconsistency} arising from locality bias and \textit{hierarchical inconsistency} caused by cumulative error propagation.
    \item We propose \textbf{AlignVAR}, a globally consistent VAR framework that enhances spatial coherence and hierarchical alignment through two complementary components: \emph{Spatial Consistency Autoregression (SCA)} and \emph{Hierarchical Consistency Constraint (HCC)}.
    \item Extensive experiments demonstrate that AlignVAR substantially improves reconstruction coherence and perceptual quality, establishing a strong benchmark for autoregressive real-world ISR.
\end{itemize}

\section{Related Works}
\label{sec:related-work}

\paragraph{Image super-resolution.}
Early image super-resolution methods learn deterministic mappings from low-resolution to high-resolution images under simplified degradations such as bicubic downsampling, limiting their generalization to real-world scenarios \cite{chen2023hat, dong2014SRCNN, liang2021swinir, lim2017edsr, zhang2018rcan, liu2025srmambat}. 
To overcome this, GAN-based approaches employ adversarial losses to approximate natural image distributions \cite{Ledig2016SRGAN, lai2017deep, wang2018esrgan, yuan2018unsupervised, zhang2019ranksrgan}. Methods like BSRGAN \cite{zhang2021bsrgan} and Real-ESRGAN \cite{wang2021realesrgan} introduce blind degradation modeling for robust real-world recovery, but often suffer from unstable training and unnatural artifacts \cite{wu2024seesr}. 
Recently, diffusion-based SR methods such as DiffBIR \cite{lin2024diffbir}, StableSR \cite{wang2024StableSR}, SeeSR \cite{wu2024seesr}, PASD \cite{yang2024pasd}, and PiSA-SR \cite{sun2025PiSASR} leverage generative priors from pretrained diffusion models to enhance perceptual fidelity. 
Despite their strong visual performance, these models rely on iterative denoising, which incurs high computational cost and may produce hallucinated or inconsistent details \cite{aithal2024hallucination, narasimhaswamy2024handiffuser}.

\vspace{-10pt}
\paragraph{Visual autoregressive models.}
Autoregression, motivated by its success in large language models~\cite{touvron2023llama, achiam2023gpt4}, has recently gained traction in visual generation tasks. 
Most approaches discretize latent features using vector quantizers~\cite{van2017VQVAE, lee2022rqvae} and predict tokens sequentially~\cite{yu2022scaling, wang2024emu3, chang2022maskgit, wei2025pure, guo2022larsr, yu2025ssc}. 
However, next-token prediction over flattened sequences often breaks spatial structure, which makes it difficult to generate coherent high-resolution content~\cite{tian2024var}. 
Visual autoregressive modeling~\cite{tian2024var} addresses this limitation through next-scale prediction, reconstructing images progressively across multiple scales and achieving strong generative performance. 
This paradigm has also been explored in image restoration~\cite{wang2024varformer} and further extended to ISR through VARSR~\cite{qu2025varsr}, which reconstructs HR images in a coarse-to-fine manner. 
Nevertheless, VARSR still suffers from locality bias and cumulative error propagation, leading to both spatial and hierarchical inconsistency.

\section{Preliminaries}
\label{sec:prelim}

In next-scale visual autoregression, image reconstruction is performed hierarchically within the latent space of a vector-quantized variational autoencoder (VQ-VAE)~\cite{van2017VQVAE}.  
At each scale $k$, the latent feature ${f}_k$ is computed as the residual between the target latent ${f}$ and the upsampled reconstructions from all previous coarser scales:
\begin{equation}
{f}_k
=
{f}
-
\sum_{m=1}^{k-1}
\mathrm{upsample}\!\big(\mathrm{lookup}(\mathcal{V}, r_m)\big),
\label{eq:var_f}
\end{equation}
where $\mathcal{V}$ denotes the shared codebook, and $\mathrm{lookup}(\mathcal{V}, r_m)$ retrieves the embedded vectors corresponding to tokens $r_m$.

Each latent position is quantized to the nearest codebook entry, yielding the discrete token map $r_k$:
\begin{equation}
r_k(i,j)
=
\arg\min_{v\in[|\mathcal{V}|]}
\big\|
\mathrm{lookup}(\mathcal{V}, v)
-
{f}_k(i,j)
\big\|_2^2.
\label{eq:var_quant}
\end{equation}

Across scales, the overall autoregressive process follows a coarse-to-fine factorization:
\begin{equation}
p(r_1, r_2, \dots, r_K)
=
\prod_{k=1}^{K}
p_\theta(r_k \mid r_{1}, r_{2}, \dots, r_{k-1}, c),
\label{eq:var_factor}
\end{equation}
where $c$ represents the conditional latent derived from the low-resolution input.

\section{Methods}
\label{sec:method}

\begin{figure*}[!t]
    \centering
    \includegraphics[width=1\linewidth]{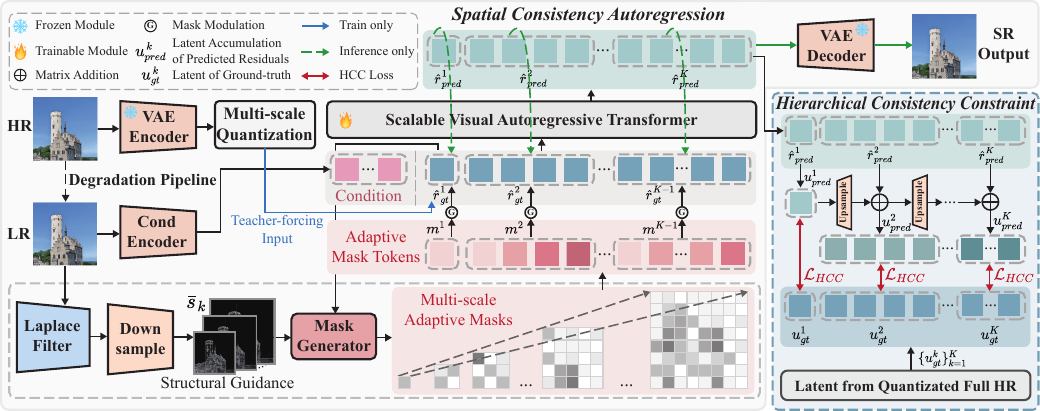}
    \vspace{-10pt}
    \caption{
        Overall architecture of the proposed \textbf{AlignVAR}.
        AlignVAR comprises two complementary components: a \textit{Spatial Consistency Autoregression (SCA)} that performs scale-wise prediction and reweights intra-scale features using adaptive masks, and a \textit{Hierarchical Consistency Constraint (HCC)} that jointly supervises residual and full representations to recalibrate inter-scale dependencies.
        }
    \vspace{-10pt}
    \label{fig:framework}
\end{figure*}

\subsection{Motivation and Empirical Observation}
\label{sec:motivation}
We analyze the spatial behavior of visual autoregression (VAR) in image super-resolution by visualizing its attention distribution.
As shown in Fig.~\ref{fig:attention}, VARSR~\cite{qu2025varsr} exhibits highly localized attention responses, 
where most attention mass is confined to narrow regions, revealing a strong \emph{local bias}.
This bias limits long-range contextual aggregation and weakens spatial correlations among distant yet related structures. 
These observations indicate that sequential tokenization and limited receptive field lead to a non-conservative spatial information flow, resulting in spatial inconsistency within each scale.
This motivates our first question:

\textbf{Q1:} \textit{How can we mitigate the local bias of VAR to achieve spatial consistency within each scale?}
\vspace{-10pt}
\paragraph{Hierarchical inconsistency caused by error accumulation.}
Beyond spatial coherence, the hierarchical prediction process of VAR inherently suffers from cumulative prediction errors, as shown in Fig.~\ref{fig:motivation-b}.
In the next-scale prediction paradigm, each latent ${r}_k$ is estimated based on imperfect outputs from preceding scales ${r}_{1:k-1}$.
Any deviation in coarse-scale predictions shifts the conditional distribution $p({r}_k \mid {r}_{1:k-1})$, leading to a compounding effect in which errors are propagated and amplified across scales.
We attribute this problem to residual-only supervision used in existing VAR frameworks, which leaves intermediate latent representations under-constrained and destabilizes the generative hierarchy.
This motivates our second question:

\textbf{Q2:} \textit{How can we recalibrate intermediate predictions to maintain hierarchical consistency across scales?}

\subsection{Overview}
Motivated by the two issues identified above, we propose \textbf{AlignVAR}, a globally consistent visual autoregressive framework for ISR. The overall architecture of AlignVAR is illustrated in Fig.~\ref{fig:framework}. Briefly, AlignVAR is composed of two primary modules: Spatial Consistency Autoregression (\textbf{SCA}) and Hierarchical Consistency Constraint (\textbf{HCC}). The \textbf{SCA} expands the effective receptive field and applies a learnable structure-aware mask to mitigate local attention bias (\S\ref{sec:sca}), while the \textbf{HCC} enforces full-scale latent alignment to recalibrate intermediate predictions and suppress cumulative drift across scales (\S\ref{sec:hrs}). 
Together, these designs enable AlignVAR to produce perceptually faithful and structurally coherent high-resolution reconstructions.

\subsection{Spatial Consistency Autoregression (SCA)}
\label{sec:sca}
The core idea of SCA is to enhance spatial coherence within the autoregressive process. 
Rather than relying on order-based attention with an inherently local bias, SCA introduces a structure-aware conditioning principle that reweights contextual dependencies.

Specifically, SCA preserves the coarse-to-fine autoregressive structure but replaces the plain context $(r_1,\dots,r_{k-1})$ with a structure-aware counterpart.
Let $\tilde{r}_k$ denote the reweighted tokens at scale $k$.
The joint distribution under SCA becomes
\begin{equation}
p_{\text{SCA}}(\tilde{r}_1, \dots, \tilde{r}_K)
=
\prod_{k=1}^{K}
p_\theta\big(\tilde{r}_k \mid \tilde{r}_{1}, \tilde{r}_{2}, \dots, \tilde{r}_{k-1}, c\big),
\label{eq:sca_factor}
\end{equation}
where the next-scale prediction at level $k$ is conditioned on structure-aligned representations of all coarser scales.
SCA thus remains defined only across scales, while spatial consistency of the context is enforced through structure-aware reweighted tokens $\tilde{r}_{1:k-1}$.

\paragraph{Structure-aware reweighting field.}
As shown in the bottom middle part of Fig.~\ref{fig:framework}, we use structure-aware guidance extracted from the low-resolution input $I_{LR}$ to capture structural cues corresponding to potential edge or texture regions:
\begin{equation}
s = \big|\mathrm{Laplacian}(I_{LR})\big|,
\qquad
\bar{s}_k = \mathrm{norm}\!\big(\mathrm{Down}_k(s)\big),
\label{eq:sca_prior}
\end{equation}
where $\mathrm{Laplacian}$ represents a Laplacian operator \cite{burt1987laplacian}, chosen for its sensitivity to second-order structural changes, enabling the guidance map to highlight edges for effective spatial reweighting. $\mathrm{Down}_k(\cdot)$ denotes scale-specific downsampling that matches the spatial resolution of the $k$-th latent map, and $\mathrm{norm}(\cdot)$ rescales values to $[0,1]$.

By normalizing across scales, the structural guidance $\bar{s}_k$ remains comparable in magnitude and semantics, allowing the model to leverage multi-scale geometric consistency rather than relying solely on local correlations.

Subsequently, we use a lightweight MLP-based mask generator $\mathcal{M}_\phi$ to predict a spatial modulation field $m_k$ from the autoregressive tokens $r_k$ and the corresponding structural guidance $\bar{s}_k$:
\begin{equation}
m_k = \sigma\!\big(\mathcal{M}_\phi([\,r_k,\, \bar{s}_k\,])\big),
\label{eq:sca_mask_head}
\end{equation}
where $[\,\cdot\,]$ denotes channel-wise concatenation and $\sigma$ is a sigmoid function.
This modulation field $m_k$ serves as a structure-aware reweighting map that adaptively adjusts the token responses according to the reliability of the underlying structure.
Regions with clear geometric cues are assigned higher weights, encouraging the model to attend to stable, well-defined features, 
while uncertain or textureless areas are softly suppressed to reduce local noise amplification.
Finally, the structure-aware reweighted tokens are produced through a learnable spatial field via token gating:
\begin{equation}
\tilde{r}_k = (1 + m_k)\odot r_k,
\label{eq:sca_reweight}
\end{equation}
with $\odot$ denoting element-wise multiplication.
This operation can be viewed as a spatially adaptive gain control that modulates the strength of autoregressive context propagation.
By reinforcing structure-aligned activations and suppressing unstable ones, 
the reweighting field guides the model to preferentially propagate information along reliable structural paths. 
This selectively amplifies long-range correlations between semantically related regions, effectively mitigating local bias and enabling broader contextual aggregation within each scale.

While SCA enforces spatial coherence within each scale, the hierarchical dependencies across scales remain unregularized.
Section~\ref{sec:hrs} introduces the Hierarchical Consistency Constraint to align inter-scale representations and suppress error propagation through the hierarchy.

\subsection{Hierarchical Consistency Constraint (HCC)}
\label{sec:hrs}

\begin{figure}[!t]
    \centering
    \includegraphics[width=1\linewidth]{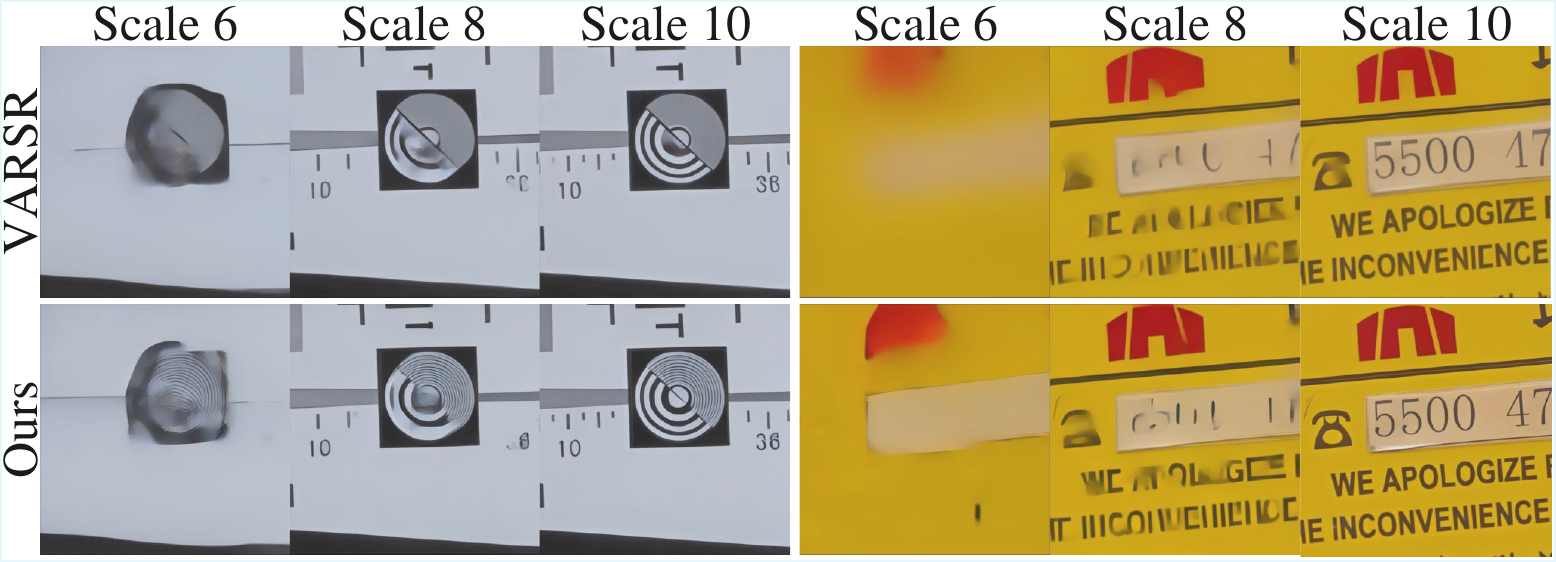}
    \caption{Comparison of multi-scale reconstructions. AlignVAR alleviates cumulative error propagation across scales and preserves more consistent structural and textural details than VARSR~\cite{qu2025varsr}.}
    \vspace{-10pt}
    \label{fig:scale}
\end{figure}
In the original next-scale autoregressive formulation, 
the Cross Entropy (CE) loss \cite{tian2024var} supervises only the residual tokens predicted at each scale. 
While effective for local refinement, this residual-only supervision cannot correct 
errors accumulated from coarser levels, allowing scale-wise biases to propagate upward. 
To address this issue, we introduce HCC, 
which provides full-scale latent supervision to explicitly recalibrate hierarchical dependencies. 
By aligning predicted latent representations with their complete multi-scale ground truths, 
HCC enforces global consistency across the entire autoregressive hierarchy.
\vspace{-5pt}
\paragraph{Full-scale representation.}
Given a high-resolution image ${I}_{HR}\in \mathbb{R}^{C\times H_K\times W_K}$, 
the VAE encoder $\mathcal{E}$ produces its latent feature representation
\begin{equation}
{z} = \mathcal{E}({I}_{HR}) \in \mathbb{R}^{C \times H_K \times W_K}.
\end{equation}
To construct scale-specific ground truths for hierarchical supervision, 
we spatially downsample ${z}$ to each target resolution $S_k$ and quantize the result into discrete tokens:
\begin{equation}
{u}_{\text{gt}}^{k} = 
\mathcal{Q}\!\big(
\mathrm{Down}({z}, S_k)
\big),
\qquad
S_k \in \{S_1, S_2, \dots, S_K\},
\label{eq:hrs_gt_latent}
\end{equation}
where $\mathrm{Down}(\cdot)$ denotes spatial downsampling to the resolution of scale $k$, and $\mathcal{Q}(\cdot)$ denotes the shared vector quantizer.
Unlike the residual tokens used in the standard CE loss, 
${u}_{\text{gt}}^{k}$ represents the complete latent state at each scale, 
thus providing a stronger supervision target.
\vspace{-5pt}
\paragraph{HCC loss.}
During autoregressive reconstruction, the model predicts residual tokens $\hat{{r}}_{\text{pred}}^{k}$ for each scale. 
The corresponding cumulative (full-scale) latent prediction is obtained by aggregating residuals from all previous scales:
\begin{equation}
\hat{{u}}_{\text{pred}}^{k} 
= 
\hat{{u}}_{\text{pred}}^{k-1} 
+ 
\hat{{r}}_{\text{pred}}^{k},
\qquad
\hat{{u}}_{\text{pred}}^{(0)} = \mathbf{0}.
\label{eq:hrs_full}
\end{equation}
We design the HCC loss to impose a multi-scale hierarchical alignment constraint between these cumulative predictions 
and the corresponding ground-truth latents:
\begin{equation}
\mathcal{L}_{\text{HCC}}
=
\sum_{k=1}^{K}
\big\|
\hat{{u}}_{\text{pred}}^{k} - {u}_{\text{gt}}^{k}
\big\|_2^2.
\label{eq:hrs_loss}
\end{equation}
This inter-scale calibration prevents error propagation and aligns the semantic content across the hierarchy.

By extending the supervision from residuals to full latent representations, 
the HCC loss bridges the semantic gap between hierarchical scales and stabilizes the coarse-to-fine refinement process.
As shown in Fig.~\ref{fig:scale}, AlignVAR reconstructs finer and more consistent textures at higher scales compared to VARSR~\cite{qu2025varsr}, 
indicating that the proposed recalibration effectively alleviates the accumulation and propagation of local errors.
Together with SCA's spatial regularization, 
it establishes a unified mechanism that preserves intra-scale coherence and inter-scale consistency 
throughout the autoregressive super-resolution process.
\subsection{Training Objective} 
Our training process is performed under the teacher-forcing paradigm \cite{tian2024var}, 
where the prediction at each scale is conditioned on the reweighted ground-truth tokens $\tilde{r}_{\mathrm{gt}}^{1:k-1}$ and the low-resolution latent feature $c$, 
providing a stable and reliable supervision path throughout the entire multi-scale prediction procedure. 
The autoregressive predictor models a categorical distribution over the codebook entries at each spatial position $(i,j)$, optimized via a standard CE objective commonly used in modern generative modeling:
\begin{equation}
\mathcal{L}_{\mathrm{CE}}
=
- \sum_{k=1}^{K}\sum_{(i,j)}
\log p_\theta
\big(
r_{\mathrm{gt}}^k(i,j)
\mid
\tilde{r}_{\mathrm{gt}}^{1:k-1},
c
\big).
\label{eq:sca_ce}
\end{equation}
Finally, the overall objective of AlignVAR combines both intra-scale and inter-scale learning signals, ensuring that each component contributes meaningfully to the reconstruction quality:
\begin{equation}
\mathcal{L}_{\mathrm{total}}
=
\mathcal{L}_{\mathrm{CE}}
+
\lambda \mathcal{L}_{\mathrm{HCC}},
\label{eq:sca_total}
\end{equation}
where $\lambda$ is a hyperparameter that balances the two components. 
Joint optimization of $\mathcal{M}_\phi$ and $p_\theta$ encourages spatially coherent predictions at each scale and hierarchically consistent dependencies across the reconstruction process, ultimately improving both stability and performance.

\begin{table*}[!t]
\centering
\caption{
Comparison with state-of-the-art methods on synthetic and real-world benchmarks. 
The best and second-best results are highlighted in \textbf{\textcolor{red}{bold red}} and \textcolor{blue}{\underline{underline blue}}, respectively.
}
\vspace{-5pt}
\label{tab:results}
\fontsize{8}{13}\selectfont
\renewcommand{\arraystretch}{0.7}
\begin{tabular}{l|c|ccc|ccccc|>{\columncolor{red!5}}c>{\columncolor{red!5}}c} 
\toprule
\multirow{2}{*}{Datasets} & \multirow{2}{*}{Metrics} & \multicolumn{3}{c|}{{GAN-based}} & \multicolumn{5}{c|}{{Diffusion-based}} & \multicolumn{2}{>{\columncolor{red!5}}c}{{VAR-based}} \\

& & {{BSRGAN}} & {{Real-ESR}} & {{SwinIR}} & {{LDM}} & {{StableSR}} & {{DiffBIR}} & {{PASD}}& {{UPSR}} & {{VARSR}} & {{AlignVAR}}\\
\midrule

\multirow{8}{*}{\centering DIV2K-Val}
 &PSNR$\uparrow$    & \second{24.42}  & 24.30  & 23.77  & 21.66  & 23.26  & 23.49  & 23.59  & \first{24.87}   & 24.41  & {24.35}\\
 &SSIM$\uparrow$    & 0.6164 & \first{0.6324} & 0.6186 & 0.4752 & 0.5670 & 0.5568 & 0.5899 & \second{0.6294}  & 0.6189 & {0.6021}\\
 &LPIPS$\downarrow$ & 0.3511 & 0.3267 & 0.3910 & 0.4887 & 0.3228 & 0.3638 & 0.3611 & 0.3173  & \second{0.2985} & \first{0.2955}\\
 &DISTS$\downarrow$ & 0.2369 & 0.2245 & 0.2291 & 0.2693 & \first{0.2116} & {0.2177} &0.2134  & \second{0.2132}  & {0.2167} & {0.2162}\\
 &FID$\downarrow$   & 50.99  & 44.34  & 44.45  & 55.04  & 28.32  & 34.55  & 39.74  & 39.48   & \second{28.64}  & \first{25.71}\\
 &MANIQA$\uparrow$  & 0.3547 & 0.3756 & 0.3411 & 0.3589 & 0.4173 & \second{0.4598} & 0.4440 & 0.3635  & 0.4137 & \first{0.4665}\\
 &CLIPIQA$\uparrow$ & 0.5253 & 0.5205 & 0.5213 & 0.5570 & \second{0.6752} & 0.6731 & 0.6573 & 0.5748  & 0.6312 & \first{0.6754}\\
 &MUSIQ$\uparrow$   & 60.18  & 59.76  & 57.21  & 57.46  &  65.19 & 65.57  & 66.58  & 62.38   & \second{66.88}  & \first{67.32}\\
\midrule

\multirow{8}{*}{\centering RealSR}
 &PSNR$\uparrow$    & \first{26.38} & 25.68 & 25.88 & 25.66 & 24.69 & 24.94 & 25.21 & 25.97 & 26.08 & \second{26.11}\\
 &SSIM$\uparrow$    & \second{0.7651} &  0.7614 & \first{0.7671} & 0.6934&  0.7090&  0.6664&  0.7140 & 0.7449 & 0.7381 & 0.7125\\
 &LPIPS$\downarrow$ & \second{0.2656} & 0.2710 & \first{0.2614} & 0.3367 & 0.3003 & 0.3485 & 0.2986 & 0.2862 & 0.2777 & 0.2871\\
 &DISTS$\downarrow$ & 0.2124  &\first{0.2060}  &\second{0.2061}  &0.2324  &0.2134  &0.2257 & 0.2125  &0.2083  &0.2111& {0.2123}\\
 &FID$\downarrow$   & 141.25 &135.14 &132.80 &133.34 &131.72 &\second{127.59} &139.42 & 140.11    &\first{118.84}& {138.81}\\
 &MANIQA$\uparrow$  & 0.3763  &0.3736  &0.3561  &0.3375  &0.4167  &0.4378  &\second{0.4418}  &0.3893  &0.4316& \first{0.4553}\\
 &CLIPIQA$\uparrow$ & 0.5114  &0.4487  &0.4433  &0.6053  &0.6200  &\second{0.6396}  &0.6009  &0.5856  &0.5953 & \first{0.6784}\\
 &MUSIQ$\uparrow$   & 63.28 &60.37 &59.28& 56.32 &65.25 &64.32 &66.61 &64.79 &\second{66.65} & \first{68.53}\\
\midrule

\multirow{8}{*}{\centering DRealSR}
 &PSNR$\uparrow$ & 28.70 & 28.61 & 28.20 & 27.78 & 27.87 & 26.57 & 27.45 & \first{29.21} & \second{28.84} & {28.54}\\
 &SSIM$\uparrow$ & \second{0.8028}  & \first{0.8052}  & 0.7983  & 0.7152  & 0.7427  & 0.6516  & 0.7539  & {0.8018}  & 0.7735 & 0.7682\\
 &LPIPS$\downarrow$ & \second{0.2858} &\first{0.2819} &0.2830 &0.3745 &0.3333 &0.4537 &0.3331 &0.2872 &0.3102 & 0.3184\\
 &DISTS$\downarrow$ &0.2144 &\second{0.2089} &0.2103 &0.2417 &0.2297 &0.2724 &0.2322& \first{0.2050} &0.2311 & 0.2340\\
 &FID$\downarrow$ &155.62  &147.66  &146.38  &164.87  &148.18  &160.67 & 173.40  & 157.83  &\first{141.99}& \second{145.44}\\
 &MANIQA$\uparrow$ &0.3441 &0.3435 &0.3311 &0.3342 &0.3897 &\second{0.4602} &0.4551& 0.3339 &0.4225 & \first{0.4685}\\
 &CLIPIQA$\uparrow$ &0.5061 &0.4525 &0.4522 &0.5984 &0.6321 &\second{0.6445} &0.6365 &0.5537 &0.6395 & \first{0.6534}\\
 &MUSIQ$\uparrow$ & 57.16  &54.27  &53.01  &51.37  &58.72  &61.06 & \second{63.69}  &55.88  &62.66 & \first{63.83}\\
\bottomrule
\end{tabular}
\end{table*}
\section{Experimental Results}
\label{sec:experiment}
\subsection{Experimental Setups}
\paragraph{Datasets.} AlignVAR is trained on LSDIR~\cite{li2023lsdir} and the first 10K face images from FFHQ~\cite{karras2019ffhq}, 
where low-resolution counterparts are synthesized using the Real-ESRGAN degradation pipeline~\cite{wang2021realesrgan}. 
For evaluation, we adopt both synthetic and real-world benchmarks. 
The synthetic validation set (DIV2K-Val) is built by randomly cropping 3K patches from DIV2K~\cite{agustsson2017DIV2K}, 
while real-world evaluation is conducted on DRealSR~\cite{wei2020drealsr} and RealSR~\cite{wang2024realsr}. 
Following~\cite{tian2024var}, all HR images are standardized to $512\times512$ and their LR inputs to $128\times128$.
\paragraph{Implementation details.} AlignVAR is trained using a scale-wise autoregressive transformer backbone consisting of 24 transformer blocks.
The model is initialized from pretrained VAR weights~\cite{tian2024var} for faster convergence. 
Training uses AdamW~\cite{adam2014adam} with a batch size of 32, weight decay of $5\times10^{-2}$, and an initial learning rate of $5\times10^{-5}$ decayed via cosine annealing. 
The model is optimized for 100 epochs, with the loss balancing coefficient $\lambda$ in Eq.~\ref{eq:sca_total} set to 1.0. 
All experiments are conducted on 8 NVIDIA H100 GPUs.

\begin{figure*}[!t]
    \centering
    \includegraphics[width=0.9\linewidth]{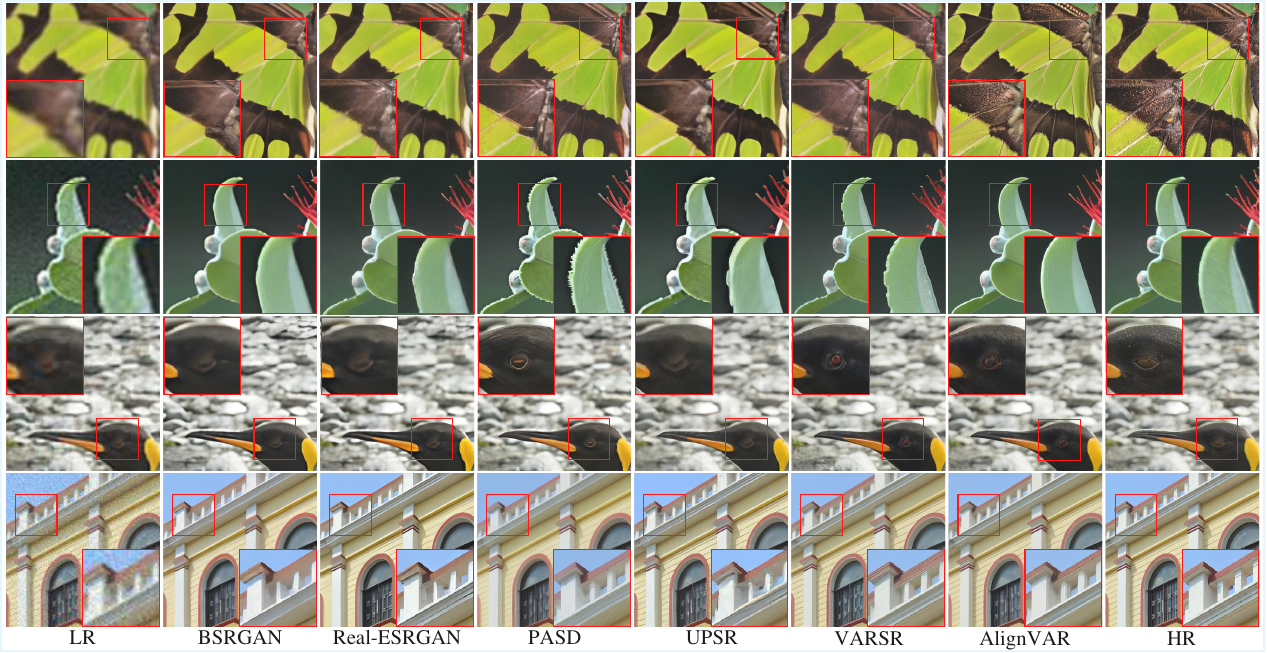}
    \vspace{-5pt}
    \caption{Qualitative comparisons with representative methods.}
    \vspace{-10pt}
    \label{fig:visual_result}
\end{figure*}
\vspace{-10pt}
\paragraph{Evaluation metrics.} To comprehensively evaluate both fidelity and perceptual quality, we adopt a diverse set of reference-based and no-reference metrics.
For fidelity, PSNR and SSIM~\cite{wang2004ssim} are computed on the Y channel of the YCbCr color space.
Perceptual similarity is measured using LPIPS~\cite{zhang2018lpips} and DISTS~\cite{ding2020dists}, while distributional alignment is assessed with FID~\cite{heusel2017fid}.
We further employ no-reference metrics, including MANIQA~\cite{yang2022maniqa}, MUSIQ~\cite{ke2021musiq}, and CLIPIQA~\cite{wang2023clipiqa}, to evaluate visual naturalness.
\vspace{-10pt}
\paragraph{Compared models.} We compare \textbf{AlignVAR} with state-of-the-art (SOTA) methods in two categories. 
The first includes GAN-based methods: BSRGAN~\cite{zhang2021bsrgan}, Real-ESRGAN~\cite{wang2021realesrgan}, and SwinIR-GAN~\cite{liang2021swinir}. 
The second covers diffusion-based methods: LDM~\cite{rombach2022ldm}, StableSR~\cite{wang2024StableSR}, DiffBIR~\cite{lin2024diffbir}, PASD~\cite{yang2024pasd}, and UPSR~\cite{zhang2025upsr}. 
We also include the recent autoregressive method VARSR~\cite{qu2025varsr}, retrained on our dataset for fairness. 
Other competing models are evaluated using official implementations and pretrained weights.

\subsection{Comparison with SOTA}
\paragraph{Quantitative comparisons.} 
As shown in Table~\ref{tab:results}, \textbf{AlignVAR} performs strongly across both synthetic and real-world benchmarks, substantially improving overall perceptual quality. 
On the DIV2K-Val dataset, it outperforms GAN-based and diffusion-based methods in all perceptual metrics, achieving the lowest FID of 25.71 and the best LPIPS of 0.2955 among all compared models. 
On RealSR, AlignVAR significantly increases MUSIQ from 66.65 to 68.53 and CLIPIQA from 0.5953 to 0.6784 compared to VARSR~\cite{qu2025varsr}, demonstrating its consistent effectiveness. 
Although AlignVAR does not achieve the highest fidelity metrics, this is expected because fine details in LR images are severely degraded and cannot be fully recovered by any existing method. AlignVAR consistently produces high-quality super-resolution results that better align with human visual perception, offering perceptually realistic, visually coherent, and structurally consistent reconstructions.
\vspace{-20pt}
\paragraph{Qualitative comparisons.} 
Fig.~\ref{fig:visual_result} presents visual comparisons with representative ISR methods. 
Some artifacts highlight common shortcomings of existing paradigms: GAN-based models often cause local distortions and jagged edges, while diffusion-based models may hallucinate textures and weaken structural alignment. 
In contrast, AlignVAR reconstructs sharp edges, coherent textures, and natural color transitions, better aligning with human visual perception. 
Notably, when true high-frequency details are irrecoverably lost in the low-resolution input, AlignVAR preserves the recoverable structures and generates perceptually plausible textures. 
This explains its clearly superior no-reference metric scores, even if the reference metric scores in Table~\ref{tab:results} are not correspondingly higher.
\begin{table}[!t]\scriptsize
    \centering
    \vspace{-5pt}
    \caption{Complexity comparison.}
    \vspace{-5pt}
    \begin{tabular}{c|c|c|c}
    \toprule
        Models &  Params & Steps & Inference Time\\
        \midrule
         StableSR & 1409.1M & 200 & 15.32s \\
         DiffBIR & 1900.4M & 20 & 5.03s \\
         PASD & 1716.7M & 50 & 5.94s \\
         UPSR & 1530.9M & \textbf{5} & 2.79s  \\
         VARSR & 1102.9M & 10 & 0.52s \\
         AlignVAR & \textbf{1056.5M} & 10 & \textbf{0.43s} \\
    \bottomrule
    \end{tabular}
    \label{tab:complexity}
\end{table}
\begin{table}[!t]\scriptsize
  \centering
  \vspace{-5pt}
  \caption{Ablation study of the SCA. 
``w/o'' denotes removing SCA, ``RI'' indicates random input, and ``SG'' represents the structural guidance used in our model.
}
\vspace{-5pt}
  \label{tab:ablation_SCA}
  \setlength{\tabcolsep}{5pt}
  \begin{tabular}{c|ccc|ccc}
    \toprule
    \multirow{2}{*}{Metrics} & \multicolumn{3}{c|}{RealSR} & \multicolumn{3}{c}{DrealSR} \\
            & w/o SCA & RI & SG & w/o SCA & RI & SG \\
    \midrule
    PSNR$\uparrow$   & \textbf{26.77} & 26.67   & 26.11  & \textbf{29.32} & 28.68 & 28.54\\
    LPIPS$\downarrow$& \textbf{0.2847} & 0.2855 & 0.2871 & \textbf{0.2975} & 0.3077 & 0.3184\\
    DISTS$\downarrow$& \textbf{0.2079} & 0.2108 & 0.2123 & \textbf{0.2253} & 0.2289 & 0.2340\\
    MANIQA$\uparrow$ & 0.4351 & 0.4435 & \textbf{0.4553} & 0.4285 & 0.4312 & \textbf{0.4685}\\
    MUSIQ$\uparrow$  & 66.74  & 67.21  & \textbf{68.53}  & 62.75  & 63.06 & \textbf{63.83}\\
    \bottomrule
  \end{tabular}
\end{table}
\vspace{-10pt}
\paragraph{Complexity comparisons.}
As shown in Table~\ref{tab:complexity}, AlignVAR achieves significantly higher efficiency than diffusion-based models and VARSR~\cite{qu2025varsr}. 
Diffusion-based methods require iterative refinement, leading to long inference times even with a limited number of sampling steps. 
In contrast, AlignVAR reconstructs a $512\times512$ image in only 0.43 seconds, which is over $10\times$ faster than PASD and more than $5\times$ faster than the 5-step UPSR. 
This efficiency stems from the scale-wise prediction strategy, where early scales involve fewer tokens and thus incur lower computational cost. 
Compared with the autoregressive baseline VARSR, which incorporates an additional diffusion refiner at the expense of increased parameters and latency, AlignVAR removes this component and adopts a lightweight mask generator that introduces negligible overhead. 
Thus, AlignVAR achieves better overall performance with fewer parameters and faster inference.
Further analysis of the efficiency advantage of AlignVAR is provided in the Supplementary Material.
\begin{figure}
    \centering
    \includegraphics[width=1\linewidth]{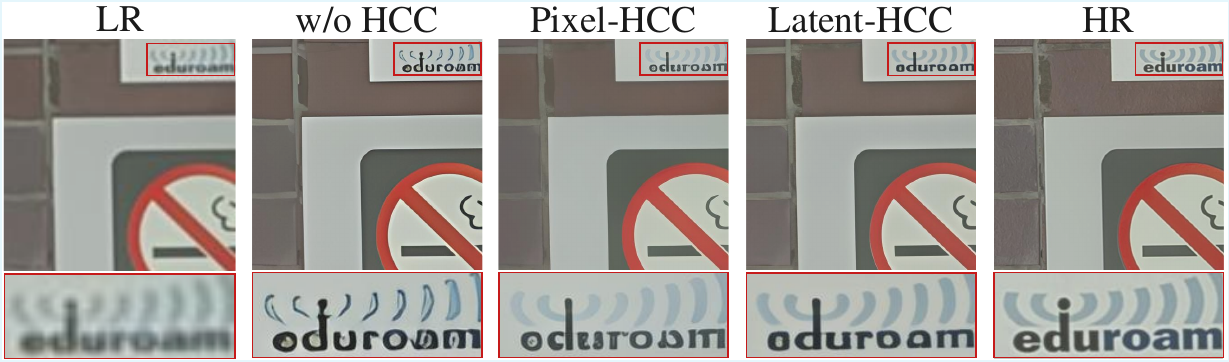}
    \caption{Comparison of supervision domains for HCC.}
    \vspace{-15pt}
    \label{fig:hcc}
\end{figure}
\subsection{Ablation Study and Analysis}
\begin{figure}[!t]
    \centering
    \includegraphics[width=1\linewidth]{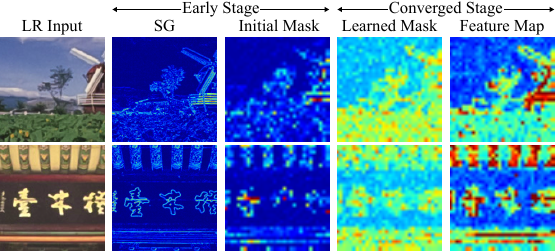}
    \caption{Analysis of the adaptive mask at different training stages. 
SG denotes the structural guidance extracted from the LR input.}
    \vspace{-10pt}
    \label{fig:mask}
\end{figure}

\paragraph{Effectiveness of SCA.} 
We evaluate the effectiveness of SCA on the RealSR and DRealSR datasets, as shown in Table~\ref{tab:ablation_SCA}. 
Three configurations are compared: removing SCA, using random input for the mask generator (RI), and applying the proposed structural guidance (SG). 
Removing SCA slightly improves fidelity metrics but noticeably degrades perceptual scores, while random input leads to unstable textures and weakened visual coherence. 
These results demonstrate that SCA effectively exploits structural guidance to enhance spatial consistency and achieves a better balance between fidelity and perceptual quality.
\vspace{-10pt}
\paragraph{Effectiveness of HCC.}
We evaluate the impact of HCC on the RealSR and DRealSR datasets, as shown in Table~\ref{tab:ablation_HCC}. 
Introducing HCC consistently improves both fidelity and perceptual metrics, indicating that aligning full-scale latent representations strengthens cross-scale information flow and stabilizes hierarchical prediction. 
We further compare applying this constraint in latent space versus in pixel space and observe that latent space supervision achieves noticeably better perceptual coherence, as illustrated in Fig.~\ref{fig:hcc}. 
Overall, HCC effectively suppresses accumulated errors and enhances inter-scale consistency, yielding reconstructions that are both more accurate and visually coherent.
\vspace{-10pt}
\paragraph{Influence of the balancing coefficient.}
We study the effect of the balancing coefficient $\lambda$ in Eq.~\ref{eq:sca_total} on the RealSR dataset, as shown in Table~\ref{tab:lambda}. 
When $\lambda$ increases from 0.5 to 1.0, both perceptual and fidelity metrics improve, with the best perceptual quality achieved at $\lambda=1.0$. 
Further increasing $\lambda$ beyond 1.0 slightly enhances distortion-oriented scores but leads to a decrease in perceptual quality. 
These results suggest that $\lambda=1.0$ provides better overall performance in terms of perceptual quality while maintaining competitive fidelity scores.
\vspace{-10pt}
\paragraph{Analysis of the Adaptive Mask.} 
Fig.~\ref{fig:mask} visualizes the structural guidance, the corresponding initial and learned masks, and the reweighted feature map. 
The structural guidance captures coarse edges and contours, which are transferred to the initial mask as a spatial bias. 
After training, the learned mask becomes sharper and more selective, concentrating on semantic boundaries such as windmill blades and Chinese characters while suppressing activations in smooth background regions. 
The reweighted feature map highlights these structural areas with clearer and more coherent responses, suggesting that the mask adaptively emphasizes perceptually important regions. 
Overall, this process enables more structure-aware feature interaction and improves spatial consistency in the reconstructed images.

\begin{figure}
    \centering
    \includegraphics[width=1\linewidth]{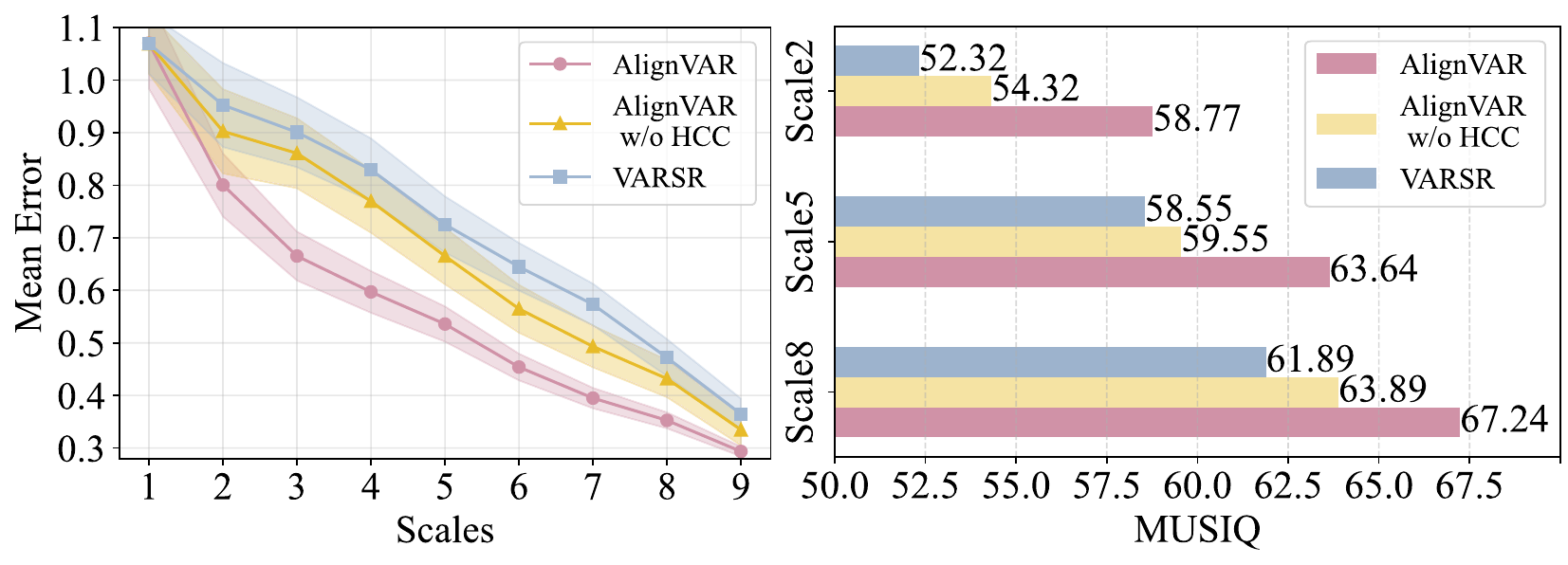}
    \vspace{-20pt}
    \caption{Quantitative analysis of the Hierarchical Consistency Constraint on RealSR. 
Shown are the mean prediction errors across scales (left) and the MUSIQ scores under identical random perturbations injected at different scales (right).}
    \vspace{-15pt}
    \label{fig:hcc-analysis}
\end{figure}

\begin{table}[!t]\scriptsize
  \centering
  \caption{Ablation on the Hierarchical Consistency Constraint.}
  \vspace{-5pt}
  \label{tab:ablation_HCC}
  \begin{tabular}{c|cc|cc}
    \toprule
    \multirow{2}{*}{Metrics} & \multicolumn{2}{c|}{RealSR} & \multicolumn{2}{c}{DrealSR} \\
            & w/o HCC & HCC & w/o HCC & HCC \\
    \midrule
    PSNR$\uparrow$   & {25.85} & \textbf{26.11} & {28.23} & \textbf{28.54} \\
    LPIPS$\downarrow$& {0.2898} & \textbf{0.2871} & {0.3208} & \textbf{0.3184} \\
    DISTS$\downarrow$& {0.2175} & \textbf{0.2123} & {0.2411} & \textbf{0.2340} \\
    MANIQA$\uparrow$ & 0.4431 & \textbf{0.4553} & 0.4442 & \textbf{0.4685} \\
    MUSIQ$\uparrow$  & 67.06  & \textbf{68.53}  & 62.85  & \textbf{63.83}  \\
    \bottomrule
  \end{tabular}
\end{table}
\begin{table}[!t]\scriptsize
    \centering
    \vspace{-5pt}
    \caption{Effect of the balancing coefficient $\lambda$ on RealSR.}
    \vspace{-5pt}
    \begin{tabular}{c|c|c|c|c|c}
        \toprule
        Settings & PSNR$\uparrow$ & LPIPS$\downarrow$ & DISTS$\downarrow$ & MANIQA$\uparrow$ & MUSIQ$\uparrow$\\
        \midrule
        $\lambda=0.5$  & 25.83 & 0.2874 & 0.2184 & 0.4403 & 67.56 \\
        $\lambda=1.0$  & 26.11 & 0.2871 & 0.2123 & \textbf{0.4553} & \textbf{68.53} \\
        $\lambda=1.5$  & 26.21 & 0.2785 & 0.2107 & 0.4465 & 67.85 \\
        $\lambda=2.0$  & \textbf{26.35} & \textbf{0.2703} & \textbf{0.2079} & 0.4337 & 67.03 \\
        \bottomrule
    \end{tabular}
    \label{tab:lambda}
\end{table}
\vspace{-10pt}
\paragraph{Analysis of HCC.} 
To further understand the effect of HCC, we analyze its impact on hierarchical recalibration and robustness to perturbations. 
Fig.~\ref{fig:hcc-analysis} (left) plots the mean squared error (MSE) at each prediction scale. 
AlignVAR without HCC already reduces overall errors compared with VARSR, and adding HCC further lowers the MSE at early and middle scales, indicating that HCC serves as a coarse-to-fine recalibration mechanism that corrects global structures before finer refinement. 
For robustness evaluation, we inject identical random perturbations into latent features at scales 2, 5, and 8 and report MUSIQ scores in Fig.~\ref{fig:hcc-analysis} (right). 
AlignVAR shows the smallest MUSIQ degradation under all noise levels, confirming that HCC improves inter-scale stability against cumulative perturbations.
\vspace{-5pt}
\section{Conclusion}
\label{sec:conclusion}
We present \textbf{AlignVAR}, a globally consistent visual autoregressive framework for image super-resolution. 
Through a detailed and comprehensive technical analysis of inconsistency issues in existing VAR models, we identify spatial local bias and hierarchical error propagation as the main bottlenecks severely hindering coherent and accurate reconstruction. 
To alleviate these issues, we introduce Spatial Consistency Autoregression to adaptively reweight fine-grained intra-scale attention and Hierarchical Consistency Constraint to recalibrate multi-level inter-scale dependencies through joint residual and full-scale supervision. 
Extensive experiments on both synthetic and real-world benchmarks consistently demonstrate that AlignVAR achieves an excellent overall balance between fidelity and perceptual quality. 
This work provides a new perspective for achieving globally coherent and reliable visual autoregression for high-quality image super-resolution.

{
    \small
    \bibliographystyle{ieeenat_fullname}
    \bibliography{main}
}
\clearpage
\setcounter{page}{1}
\setcounter{section}{0}
\maketitlesupplementary

\noindent This supplementary material provides additional analyses and results that extend the findings in the main paper. To clarify the procedural details of our method, \S Sec.~\ref{sec:train_infer} presents the complete training and inference pipelines of AlignVAR. To assess computational efficiency, \S Sec.~\ref{sec:complexity} provides both theoretical and empirical complexity analyses. To further verify perceptual quality, \S Sec.~\ref{sec:userstudy} reports the results of a user study comparing AlignVAR with several state-of-the-art baselines. To evaluate real-world robustness, \S Sec.~\ref{sec:reallr200} includes additional experiments on the challenging RealLR200 dataset. To better understand the behavior of our model, \S Sec.~\ref{sec:attn_vis} visualizes multi-scale attention maps produced by AlignVAR. To examine the contribution of each proposed component, \S Sec.~\ref{sec:ablation} conducts an ablation analysis of SCA and HCC using the Edge IoU metric. To investigate how SCA affects the receptive range, \S Sec.~\ref{sec:sca_range} compares attention maps with and without SCA. To analyze the influence of the hyperparameter $\lambda$, \S Sec.~\ref{sec:hyperparameter} studies its effect on loss convergence and prediction accuracy across scales. To further showcase visual improvements, \S Sec.~\ref{sec:vis} presents extended qualitative comparisons. Finally, \S Sec.~\ref{sec:limit} discusses the limitations of AlignVAR and outlines possible directions for future work.

\section{Training and Inference Procedures}
\label{sec:train_infer}
This section provides the procedural details of AlignVAR during both training and inference. 
All notations follow those introduced in the main paper. 
The overall training pipeline is summarized in Algorithm~\ref{alg:train_alignvar}, 
and the autoregressive reconstruction process is presented in Algorithm~\ref{alg:infer_alignvar}.

\paragraph{Training procedure.}
During training, AlignVAR follows a scale-wise teacher-forcing strategy, as outlined in 
Algorithm~\ref{alg:train_alignvar}. 
Given an HR--LR pair, the frozen VAE encoder \cite{van2017VQVAE} extracts the full-resolution latent 
$z \in \mathbb{R}^{C \times H_K \times W_K}$, 
where $C$ denotes the channel dimension and $(H_K, W_K)$ is the resolution of the last scale. 
For each scale $k$, the corresponding ground-truth latent 
$u_k^{\mathrm{gt}} \in \{1,\dots,|V|\}^{H_k \times W_k}$ 
is obtained via spatial downsampling and quantization, where $(H_k, W_k)$ denote the height and width of scale $k$, and $|V|$ denotes the size of the codebook. 
The residual token is defined as 
$r^k_{\mathrm{gt}} = u^k_{\mathrm{gt}} - u^{k-1}_{\mathrm{gt}}$. 
The autoregressive predictor $p_\theta$ receives the reweighted context 
$\tilde{r}^{1:k-1}_{\mathrm{gt}}$, where the modulation mask 
$m_k \in \mathbb{R}^{1 \times H_k \times W_k}$ 
is produced by the mask generator $M_\phi$ using the structural guidance extracted from the LR input. 
The model predicts the categorical distribution over the $|V|$ codebook entries for $r_k$, and the parameters of $p_\theta$ and $M_\phi$ are jointly optimized using the multi-scale cross-entropy loss $\mathcal{L}_{\mathrm{CE}}$ and the hierarchical consistency loss $\mathcal{L}_{\mathrm{HCC}}$ applied to the cumulative predictions.

\begin{algorithm}[t]
\caption{AlignVAR Training}
\label{alg:train_alignvar}
\begin{algorithmic}[1]
\State \textbf{Inputs:} HR--LR image pair $(I_{\mathrm{HR}}, I_{\mathrm{LR}})$
\State \textbf{Hyperparams:} steps $K$, resolutions $\{(H_k, W_k)\}_{k=1}^K$
\State $z = \mathcal{E}(I_{\mathrm{HR}})$; $c = \mathcal{E}_{con}(I_{\mathrm{LR}})$; $s = |\mathrm{Laplacian}(I_{\mathrm{LR}})|$;
\State $u_{\mathrm{gt}}^0 = 0$, \, ${u}_{\mathrm{pred}}^0 = 0$;
\For{$k = 1,\dots,K$}
    \State $u^k_{\mathrm{gt}} = \mathrm{Quant}(\mathrm{Down}(z, H_k, W_k))$;
    \State $r^k_{\mathrm{gt}} = u^k_{\mathrm{gt}} - u^{k-1}_{\mathrm{gt}}$;
\EndFor
\For{$k = 1,\dots,K$}
    \State $\bar{s}_k = \mathrm{norm}(\mathrm{Down}_k(s))$;
    \State $m_k = \sigma(M_\phi([\,r^k_{\mathrm{gt}},\,\bar{s}_k\,]))$;
    \State $\tilde r^k_{\mathrm{gt}} = (1+m_k)\odot r^k_{\mathrm{gt}}$;
    \State Predict $p_\theta(\tilde r_k \mid \tilde r^{1:k-1}_{\mathrm{gt}}, c)$;
    \State ${u}^k_{\mathrm{pred}} = {u}^{k-1}_{\mathrm{pred}} + \hat{r}^k_{\mathrm{pred}}$;
\EndFor
\State Compute $\mathcal{L}_{\mathrm{CE}}$ and $\mathcal{L}_{\mathrm{HCC}}$;
\State Update $(\theta, \phi)$;
\State
\Return trained model parameters;
\end{algorithmic}
\end{algorithm}

\paragraph{Inference procedure.}
During inference, the model performs fully autoregressive multi-scale prediction, as detailed in 
Algorithm~\ref{alg:infer_alignvar}.  
The LR input is first encoded into the conditional latent 
$c \in \mathbb{R}^{C \times H_K \times W_K}$, 
and the structural guidance map is computed in the same way as during training. 
Starting from ${u}^0_{\mathrm{pred}} = 0$, the model sequentially predicts the residual tokens 
$\hat{r}_k \in \{1,\dots,|V|\}^{H_k \times W_k}$, generates the mask 
$m_k$, computes the reweighted token map 
$\tilde{r}_k$, and updates the cumulative latent state 
${u}^k_{\mathrm{pred}}$. 
After all scales are completed, the final latent is mapped through the shared VQ codebook and decoded by the VAE decoder to produce the super-resolved output.

\begin{algorithm}[t]
\caption{AlignVAR Inference}
\label{alg:infer_alignvar}
\begin{algorithmic}[1]
\State \textbf{Inputs:} LR image $I_{\mathrm{LR}}$
\State \textbf{Hyperparams:} steps $K$, resolutions $\{(H_k, W_k)\}_{k=1}^K$
\State $c = \mathcal{E}_{con}(I_{\mathrm{LR}})$; \quad $s = |\mathrm{Laplacian}(I_{\mathrm{LR}})|$;
\State ${u}_{\mathrm{pred}}^0 = 0$;
\For{$k = 1,\dots,K$}
    \State $\bar{s}_k = \mathrm{norm}(\mathrm{Down}_k(s))$;
    \State $\hat{r}_k = \arg\max p_\theta(\tilde r_k \mid \tilde r_{1:k-1}, c)$;
    \State $m_k = \sigma(M_\phi([\,\hat{r}_k,\,\bar{s}_k\,]))$;
    \State $\tilde r_k = (1+m_k)\odot \hat{r}_k$;
    \State ${u}^k_{\mathrm{pred}} = {u}^{k-1}_{\mathrm{pred}} + \hat{r}_k$;
\EndFor
\State ${f}_K = \mathrm{lookup}(V,\,{u}^K_{\mathrm{pred}})$;
\State ${I}_{\mathrm{SR}} = D({f}_K)$;
\State
\Return reconstructed SR image ${I}_{\mathrm{SR}}$;
\end{algorithmic}
\end{algorithm}

\section{Complexity Analysis}
\label{sec:complexity}
\paragraph{Theoretical complexity.}
We analyze the computational cost of AlignVAR by examining its multi-scale autoregressive generation process. Let the latent resolution sequence be $\{(h_1, w_1), (h_2, w_2), \ldots, (h_K, w_K)\},$
where $(h_k, w_k)$ denotes the height and width of the VQ code map at the $k$-th autoregressive step, and the final resolution satisfies $h_K = h,\; w_K = w.$
For simplicity, we assume $n_k = h_k = w_k$ for all scales. Following the progressive-resolution design \cite{tian2024var}, we set $n_k = a^{k-1}$ with a constant $a > 1$, chosen such that $a^{K-1} = n.$

At scale $k$, the model attends to all accumulated token maps $(r_1, r_2, \ldots, r_k),$
and the total number of tokens is
\begin{equation}
\sum_{i=1}^{k} n_i^2 
    = \sum_{i=1}^{k} a^{2(i-1)} 
    = \frac{a^{2k}-1}{a^{2}-1}.
\end{equation}
Thus, the computational cost of the $k$-th autoregressive step is
\begin{equation}
\left( \frac{a^{2k}-1}{a^{2}-1} \right)^{2}.
\end{equation}

Summing over all steps yields the total generation complexity:
\begin{equation}\scriptsize
\begin{aligned}
&\sum_{k=1}^{\log_a(n)+1}
\left( \frac{a^{2k}-1}{a^{2}-1} \right)^{2} \\
&=
\frac{
(a^{4}-1)\log n 
+ \left(a^{8}n^{4}-2a^{6}n^{2}-2a^{4}(n^{2}-1)+2a^{2}-1\right)\log a
}{
(a^{2}-1)^{3}(a^{2}+1)\log a
} \\
&\sim \mathcal{O}(n^{4}).
\end{aligned}
\end{equation}

Since the final resolution satisfies $a^{K-1} = n,$ the total complexity is dominated by the last autoregressive step, yielding an overall complexity of $O(n^{4}).$

Next, we consider the additional components introduced by AlignVAR, including the structural-guidance mask generator $M_\phi$ and the residual modulation operation. The mask map at scale $k$ is denoted by $m_k \in \mathbb{R}^{1 \times n_k \times n_k}.$
Computing $m_k$ requires a single forward pass over the feature map, followed by element-wise modulation, both of which scale linearly with $n_k^{2}.$ Since this is asymptotically smaller than the autoregressive cost 
$\left(\sum_{i=1}^{k} n_i^2\right)^{2},$ 
the additional modules do not influence the overall complexity.

In summary, the theoretical generation complexity of AlignVAR remains $O(n^{4}),$ showing that the proposed consistency-enhancing modules maintain the same asymptotic computational cost.

\begin{figure}[t]
    \centering
    \includegraphics[width=0.95\linewidth]{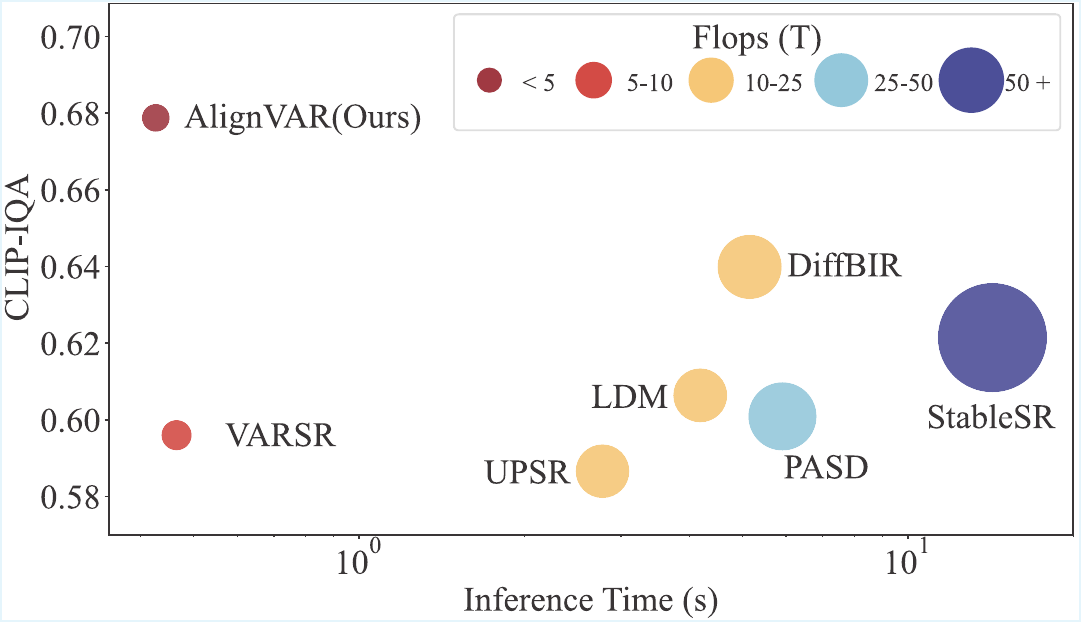}
    \caption{%
    Computational comparison among diffusion-based SR methods (DiffBIR~\cite{lin2024diffbir}, LDM~\cite{rombach2022ldm}, PASD~\cite{yang2024pasd}, UPSR~\cite{zhang2025upsr}), the VAR-based baseline (VARSR~\cite{qu2025varsr}), and our AlignVAR. AlignVAR achieves high performance with substantially lower complexity than other models.
    }
    \label{fig:complexity_bubble}
\end{figure}

\paragraph{Empirical comparison.}
To further evaluate the practical complexity, we compare AlignVAR with representative diffusion-based SR methods and the VAR-based baseline VARSR. 
We measure inference time, FLOPs, and perceptual quality (CLIP-IQA \cite{wang2023clipiqa}) under the same hardware and input configuration. 
As shown in Fig.~\ref{fig:complexity_bubble}, AlignVAR achieves the best accuracy–efficiency balance: it delivers significantly higher CLIP-IQA than diffusion models while maintaining much lower FLOPs and substantially faster inference. 
Compared with VARSR, AlignVAR attains a large perceptual gain while introducing negligible computational overhead, consistent with our theoretical analysis. 
These results demonstrate that AlignVAR retains the computational advantages of VAR while notably enhancing global consistency and reconstruction quality.

\begin{table*}[!t]
\centering
\caption{
Comparison with state-of-the-art methods on RealLR200. 
The best and second-best results are highlighted in \textbf{\textcolor{red}{bold red}} and \textcolor{blue}{\underline{underline blue}}, respectively.
}
\vspace{-5pt}
\label{tab:reallr200}
\fontsize{8}{13}\selectfont
\renewcommand{\arraystretch}{0.8}
\begin{tabular}{l|c|ccc|ccccc|>{\columncolor{red!5}}c>{\columncolor{red!5}}c} 
\toprule
\multirow{2}{*}{Datasets} & \multirow{2}{*}{Metrics} & \multicolumn{3}{c|}{{GAN-based}} & \multicolumn{5}{c|}{{Diffusion-based}} & \multicolumn{2}{>{\columncolor{red!5}}c}{{VAR-based}} \\

& & {{BSRGAN}} & {{Real-ESR}} & {{SwinIR}} & {{LDM}} & {{StableSR}} & {{DiffBIR}} & {{PASD}}& {{UPSR}} & {{VARSR}} & {{AlignVAR}}\\
\midrule

\multirow{3}{*}{\centering RealLR200}
 &NIQE$\downarrow$   & 4.3817  & 4.2048  & 4.2157  & 4.2533  & 4.2516  & \second{4.1715}  & 4.9330  & 4.7606   & 4.4579  & \first{4.0617}\\
 &MANIQA$\uparrow$  & 0.5462 & 0.5582 & 0.3741 & 0.3049 & 0.5841 & \second{0.6066} & 0.5902 & 0.4206  & 0.4536 &  \first{0.6237}\\
 &CLIPIQA$\uparrow$ & 0.5679 & 0.5389 & 0.5596 & 0.5253 & 0.6068 & \first{0.6797} & 0.6509 & 0.6397  & 0.6144 & \second{0.6734}\\
 &MUSIQ$\uparrow$   & 64.87  & 62.94  & 63.55  & 55.19  &  63.30 & \second{68.20}  & 62.06  & 66.46   & 62.12  & \first{69.36}\\

\bottomrule
\end{tabular}
\end{table*}

\begin{figure}[t]
    \centering
    \includegraphics[width=0.75\linewidth]{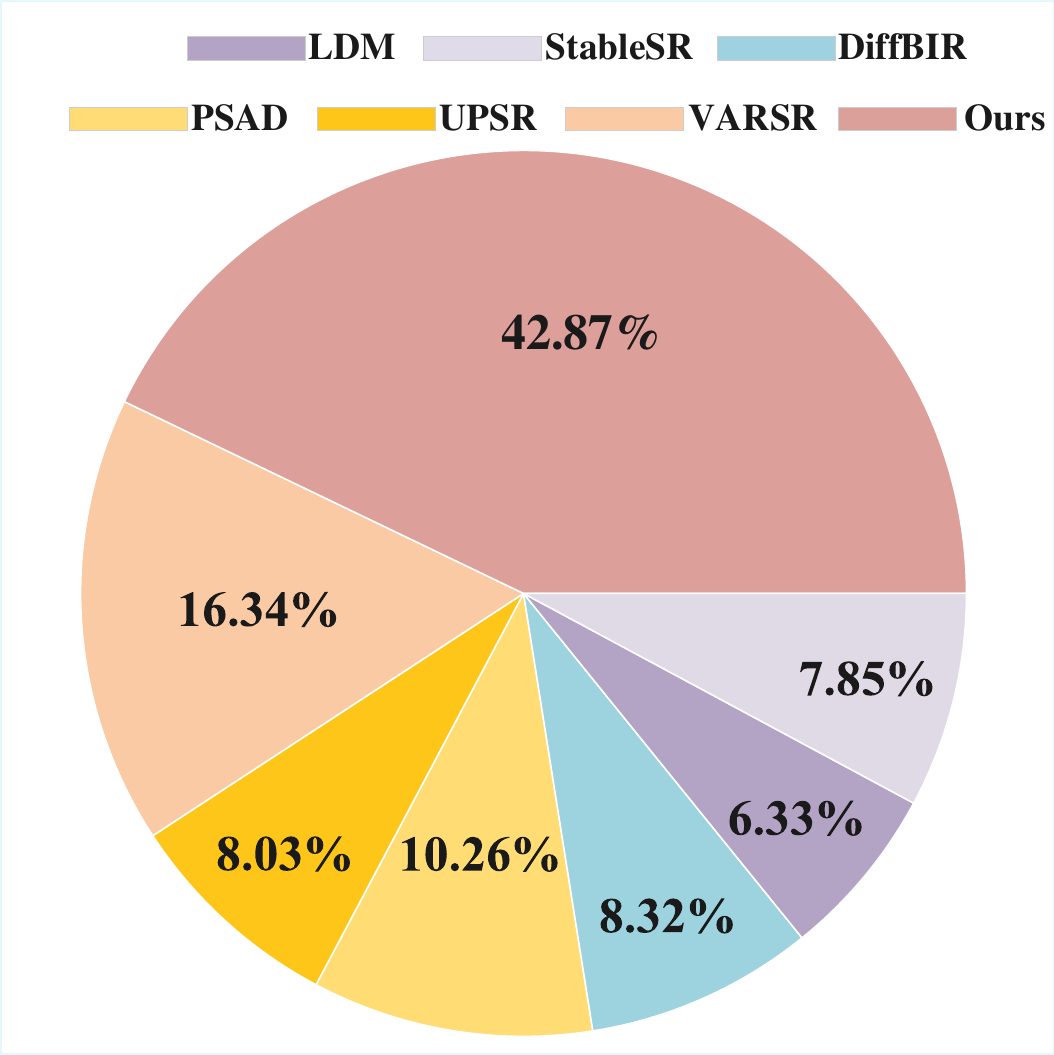}
    \caption{User study results comparing the super-resolution outputs of different models.}
    \label{fig:user_study}
\end{figure}

\section{User Study}
\label{sec:userstudy}
To evaluate the perceptual quality of the super-resolution results generated by AlignVAR compared to other methods, we conducted a user study. We randomly selected 50 images from three different datasets: RealSR~\cite{wang2024realsr}, DRealSR~\cite{wei2020drealsr}, and DIV2K-val~\cite{agustsson2017DIV2K}. These images were then presented to 20 participants, who were asked to compare the super-resolution results produced by AlignVAR against those from several competing models. The models evaluated in the study included VARSR~\cite{qu2025varsr}, PASD~\cite{yang2024pasd}, DiffBIR~\cite{lin2024diffbir}, UPSR~\cite{zhang2025upsr}, StableSR~\cite{wang2024StableSR}, and LDM~\cite{rombach2022ldm}. Participants were asked to choose which model's output they preferred for each image. The results of the user study are shown in Fig.~\ref{fig:user_study}, which highlight the superiority of AlignVAR in producing high-quality reconstructed images as judged by human evaluators, further validating the effectiveness of the proposed method.

\section{Evaluation on the RealLR200 Dataset}
\label{sec:reallr200}
To further assess the robustness and real-world applicability of AlignVAR, we conduct additional experiments on the RealLR200~\cite{wu2024seesr} dataset, a challenging collection of real-world low-resolution images with diverse and complex degradations. We compare AlignVAR against representative GAN-based methods (BSRGAN~\cite{zhang2021bsrgan}, Real-ESR~\cite{wang2021realesrgan}, SwinIR~\cite{liang2021swinir}), diffusion-based approaches (LDM~\cite{rombach2022ldm}, StableSR~\cite{wang2024StableSR}, DiffBIR~\cite{lin2024diffbir}, PASD~\cite{yang2024pasd}, UPSR~\cite{zhang2025upsr}), and the VAR-based baseline VARSR~\cite{qu2025varsr}. Both quantitative and qualitative comparisons are performed.

Table~\ref{tab:reallr200} summarizes the no-reference quality evaluation across four widely used perceptual metrics: NIQE~\cite{mittal2012niqe}, MANIQA~\cite{yang2022maniqa}, CLIP-IQA~\cite{wang2023clipiqa}, and MUSIQ~\cite{ke2021musiq}. AlignVAR consistently achieves the best or second-best performance among all competing methods. Notably, AlignVAR surpasses the diffusion-based models, which typically excel in perceptual realism, demonstrating a superior balance between fidelity, sharpness, and naturalness. Compared with VARSR~\cite{qu2025varsr}, AlignVAR obtains substantial improvements across all metrics, highlighting the effectiveness of the proposed consistency-enhancing mechanisms.

\begin{figure*}[h]
    \centering
    \includegraphics[width=0.85\linewidth]{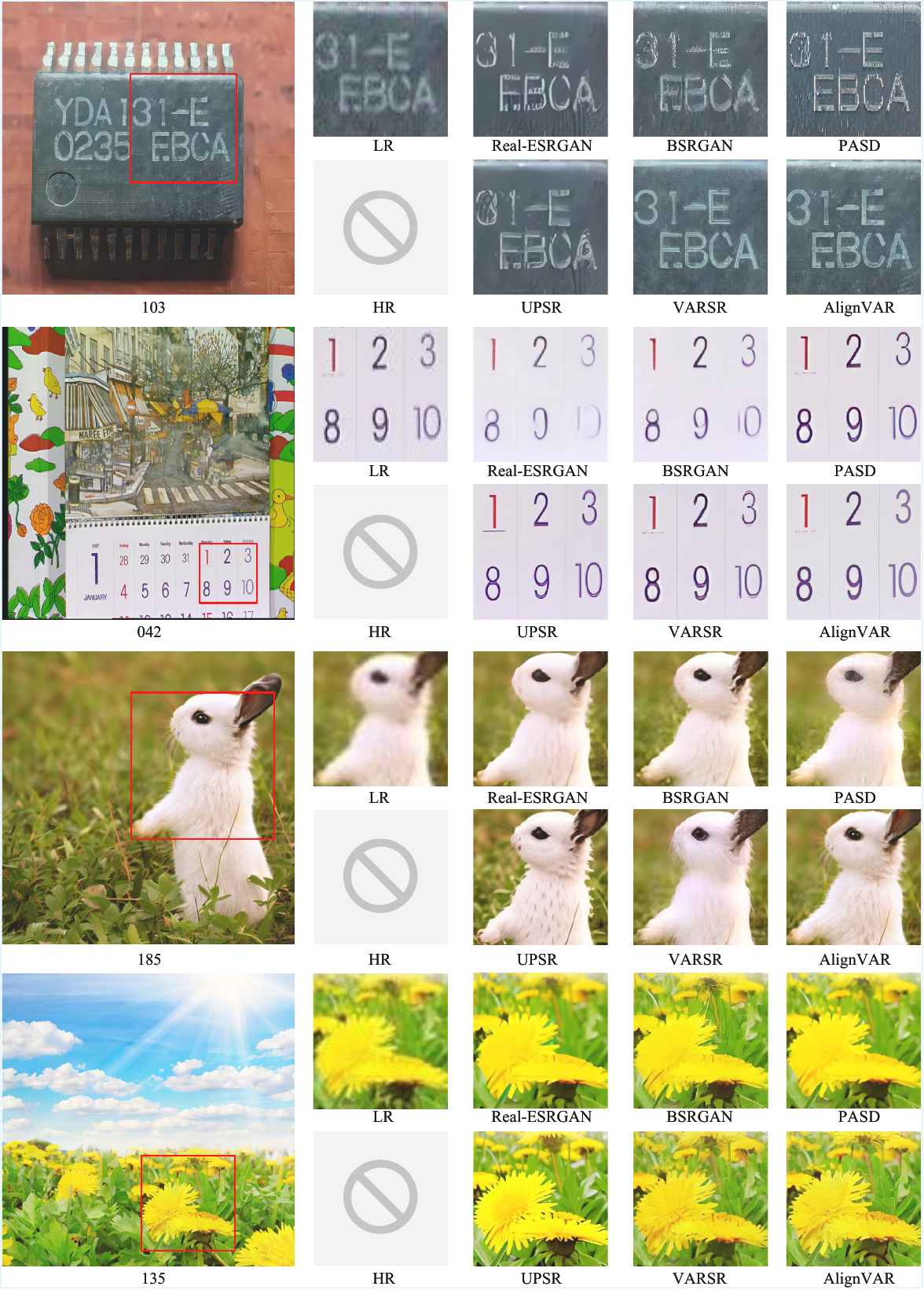}
    \caption{
        Comparison with state-of-the-art methods on the RealLR200 dataset. 
        The large image on the left is the output of AlignVAR, while the small patches on the right show enlarged crops from the LR input and competing models.
    }
    \label{fig:reallr200}
\end{figure*}

Qualitative results on RealLR200 are presented in Fig.~\ref{fig:reallr200}. GAN-based models introduce unnatural textures or hallucinated structures, while diffusion-based methods may produce inconsistent details under strong degradations. In contrast, AlignVAR reconstructs coherent edges, stable textures, and visually pleasing details. The improvements are evident in object boundaries and repeated patterns, where spatial and hierarchical consistency play a crucial role.

\section{Attention Map Visualization}
\label{sec:attn_vis}
\begin{figure}[t]
    \centering
    \includegraphics[width=1\linewidth]{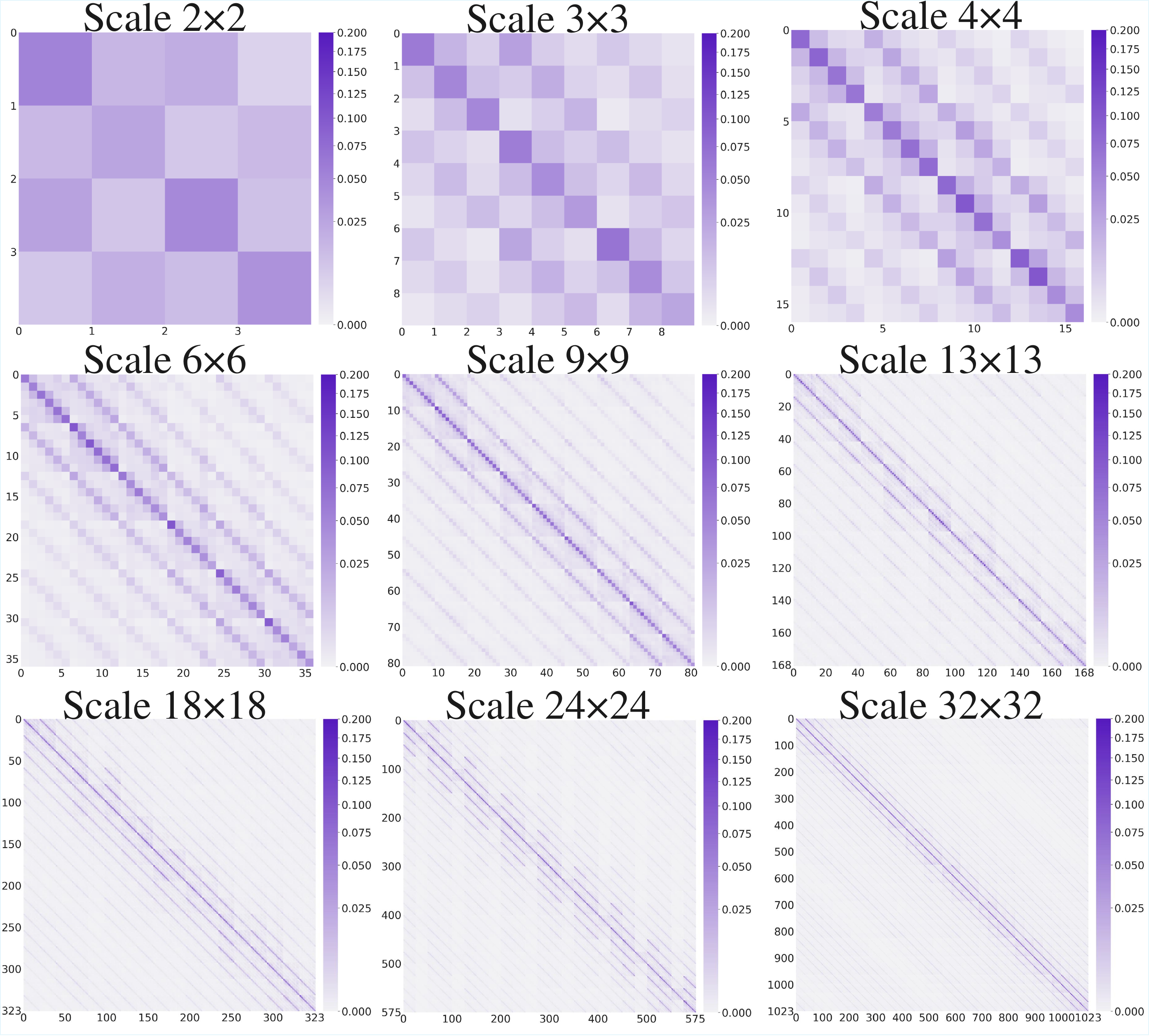}
    \caption{Attention maps at different scales. AlignVAR expands the attention range, capturing dependencies over large areas.}
    \label{fig:attention_maps}
\end{figure}

To further demonstrate the effectiveness of our method in expanding the attention range and enhancing the reweighting process, we visualize the attention maps at various scales. Specifically, we display the attention maps for all scales except for the first one, as shown in Fig.~\ref{fig:attention_maps}. From the visualizations, it is evident that our approach does not limit the attention to the diagonal elements but instead extends the attention across the entire spatial domain at all scales. This highlights the ability of our model to effectively reweight and capture dependencies over larger regions as the resolution increases. The expansive attention observed at higher scales is a clear indication that the reweighting mechanism in AlignVAR facilitates a more global understanding of the image structure.

\section{Ablation Study on SCA and HCC}
\label{sec:ablation}
\begin{figure}[t]
    \centering
    \includegraphics[width=0.8\linewidth]{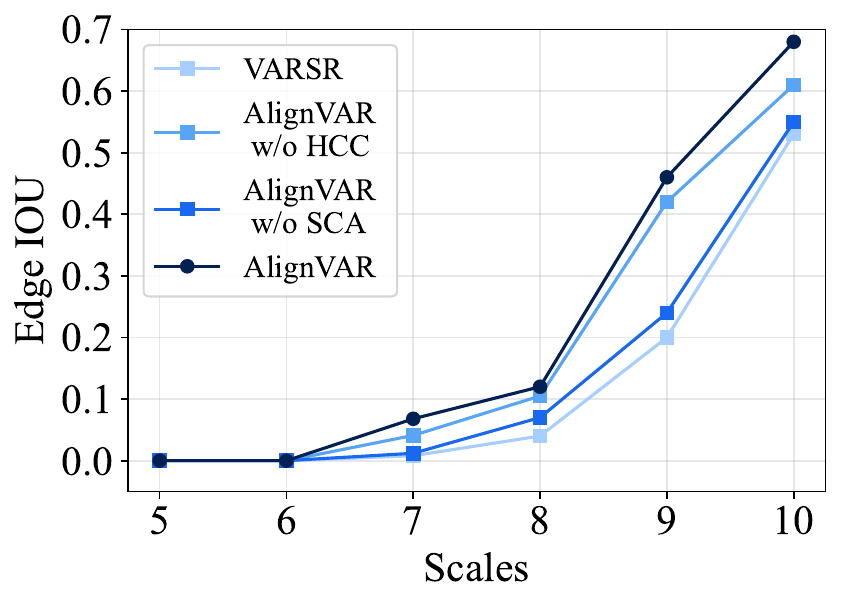}
    \vspace{-10pt}
    \caption{
        Ablation results of SCA and HCC evaluated using Edge IoU \cite{yu2016IoU} across different scales. 
        The full AlignVAR achieves the highest edge alignment quality at all scales. 
        Removing SCA results in the largest degradation, demonstrating its key role in maintaining spatial consistency and recovering fine-grained edges. 
        HCC provides complementary improvements by enhancing hierarchical consistency, but its influence is comparatively weaker.
    }
    \label{fig:ablation}
\end{figure}

We perform an ablation study to assess the individual and combined impact of SCA and HCC on the performance of AlignVAR. Specifically, we evaluate the Edge Intersection over Union (Edge IoU) \cite{yu2016IoU} across different scales, computed between the predicted edge map and the ground-truth edge map. The edge maps are extracted using the Canny edge detector, and the IoU is then calculated on the resulting binary edge masks to quantify the accuracy of structural recovery. Fig.~\ref{fig:ablation} presents the results of the ablation study. The full AlignVAR model, which incorporates both SCA and HCC, consistently achieves the highest Edge IoU. As the scale increases, the benefit becomes more apparent, demonstrating the importance of enforcing multi-scale consistency in high-resolution reconstruction. When comparing the variants, removing SCA leads to the most significant performance drop. This indicates that SCA plays a primary role in expanding the spatial attention range and maintaining spatial consistency, which is essential for recovering fine-grained and structurally aligned edges. Removing HCC also results in a decline in Edge IoU, though the degradation is less severe. This suggests that HCC mainly contributes to stabilizing cross-scale dependencies, but its influence on edge localization is comparatively weaker than that of SCA.

\section{Impact of SCA on Attention Range}
\label{sec:sca_range}
\begin{figure}[!t]
    \centering
    \includegraphics[width=1\linewidth]{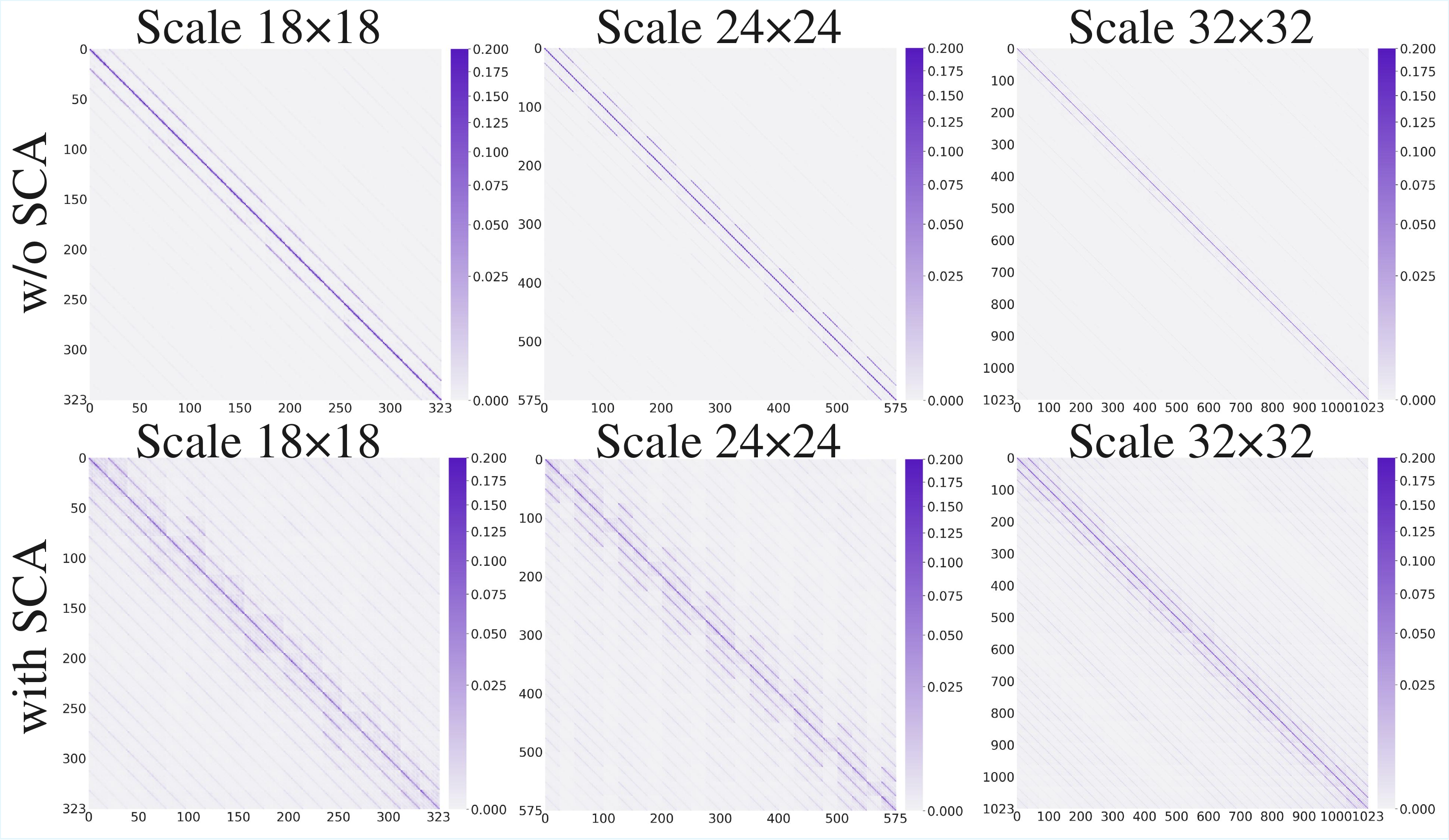}
    \caption{Impact of Spatial Consistency Autoregression (SCA) on attention maps. The attention range is significantly broader when SCA is applied.}
    \vspace{-20pt}
    \label{fig:sca_ablation}
\end{figure}

To assess the influence of the SCA on the attention range, we conduct an ablation study by comparing the attention maps with and without the inclusion of SCA. The results are presented in Fig.~\ref{fig:sca_ablation}. Without SCA, the attention maps at different scales demonstrate a more localized focus, with attention primarily concentrated along the diagonal. This suggests that, in the absence of SCA, the model tends to focus on nearby regions, limiting its ability to capture long-range dependencies across the spatial domain. In contrast, when SCA is incorporated, the attention maps exhibit a noticeable expansion. The model begins to attend to a much broader area beyond the diagonal, with significant attention given to regions that are further apart. This comparison demonstrates that SCA significantly increases the attention range, allowing the model to better capture both local and global dependencies, which contributes to the improved performance of AlignVAR.

\section{Effect of Hyperparameter}
\label{sec:hyperparameter}
\begin{figure*}[!t]
    \centering
    \includegraphics[width=1\linewidth]{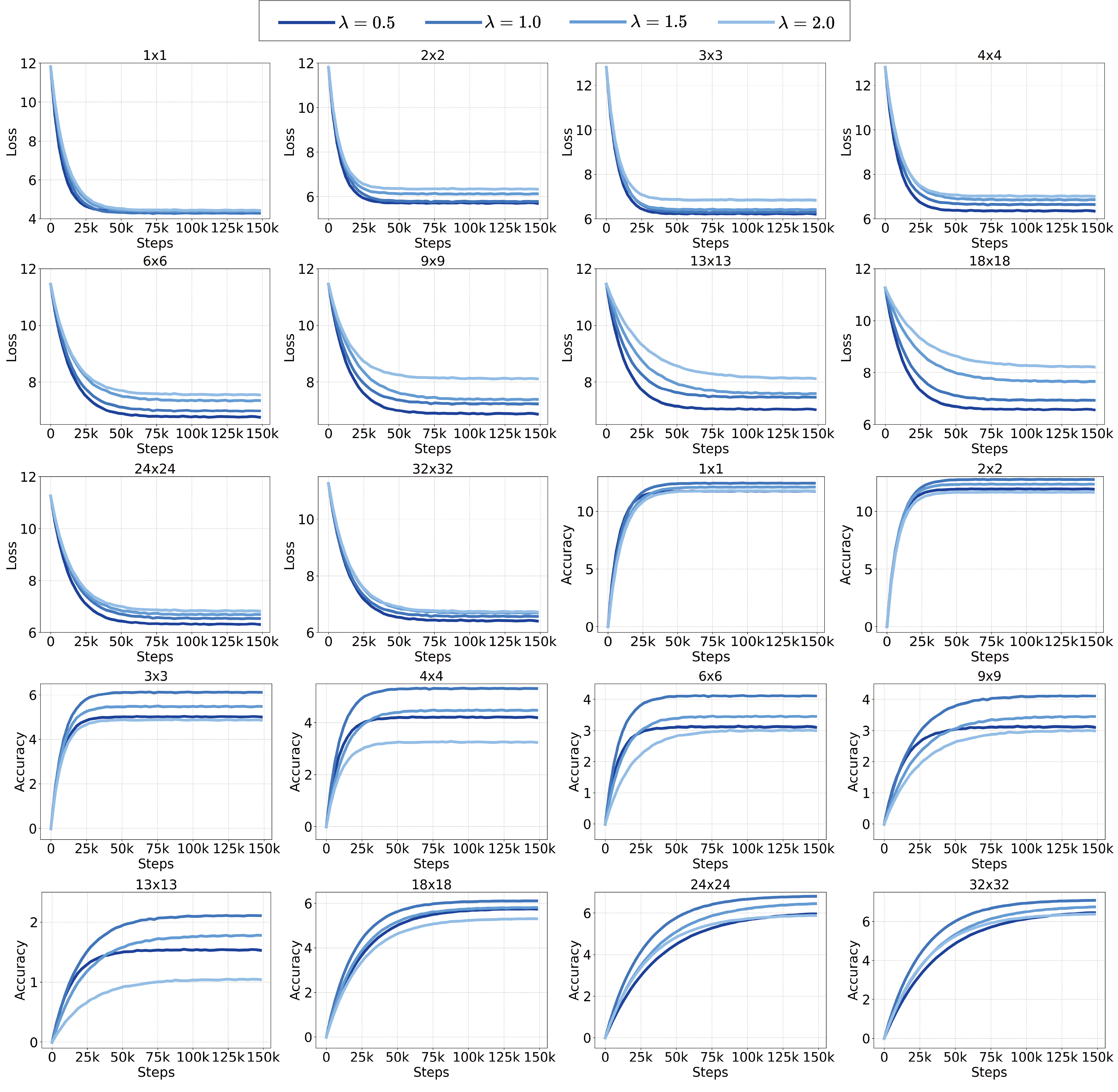}
    \caption{
        Effect of the hyperparameter $\lambda$ on training dynamics. 
        We compare four values of $\lambda$ (0.5, 1.0, 1.5, 2.0) in terms of loss and token prediction accuracy across different scales. 
        A smaller $\lambda$ leads to faster convergence but lower accuracy, particularly at the middle scales (e.g., $6 \times 6$ and $9 \times 9$), while $\lambda = 1.0$ achieves the best overall accuracy with stable convergence. 
        Larger $\lambda$ values slow down loss convergence and degrade accuracy, indicating that excessively large consistency constraints hinder optimal learning.
    }
    \label{fig:lambda_effect}
\end{figure*}

We investigate the effect of the hyperparameter $\lambda$ on the model's performance. Specifically, we analyze the impact of four different values of $\lambda$: 0.5, 1.0, 1.5, and 2.0, on the loss and accuracy across different scales. The results are shown in Fig.~\ref{fig:lambda_effect}. As observed, when $\lambda$ is small (e.g., 0.5), the model converges faster, but the token prediction accuracy is lower, particularly for the middle scales such as $6 \times 6$ and $9 \times 9$. In contrast, when $\lambda$ is set to 1.0, the model achieves the highest accuracy, with a more fast convergence. As $\lambda$ increases further (1.5 and 2.0), the loss converges more slowly, and the accuracy starts to decrease, suggesting that higher values of $\lambda$ may hinder the model’s ability to reach optimal performance for these scales. In summary, the value of $\lambda$ plays a crucial role in balancing convergence speed and accuracy. A $\lambda$ value of 1.0 provides the best trade-off between fast convergence and high accuracy.

\begin{figure*}[h]
    \centering
    \includegraphics[width=0.85\linewidth]{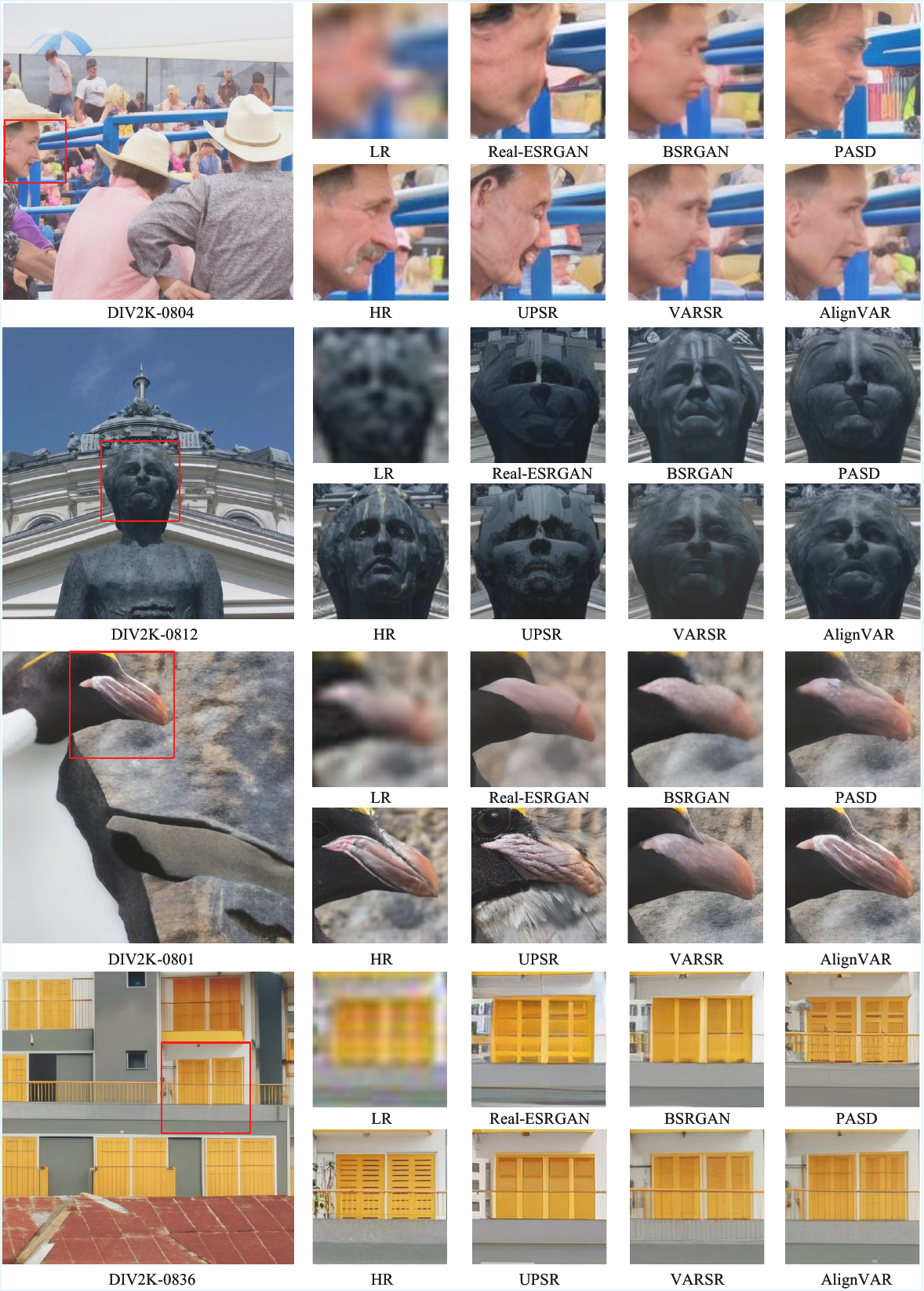}
    \caption{Additional qualitative comparisons. The large image on the left shows the output of AlignVAR, while the small patches on the right present enlarged crops from the LR input and competing methods.}
    \label{fig:visual_results1}
\end{figure*}
\begin{figure*}[h]
    \centering
    \includegraphics[width=0.85\linewidth]{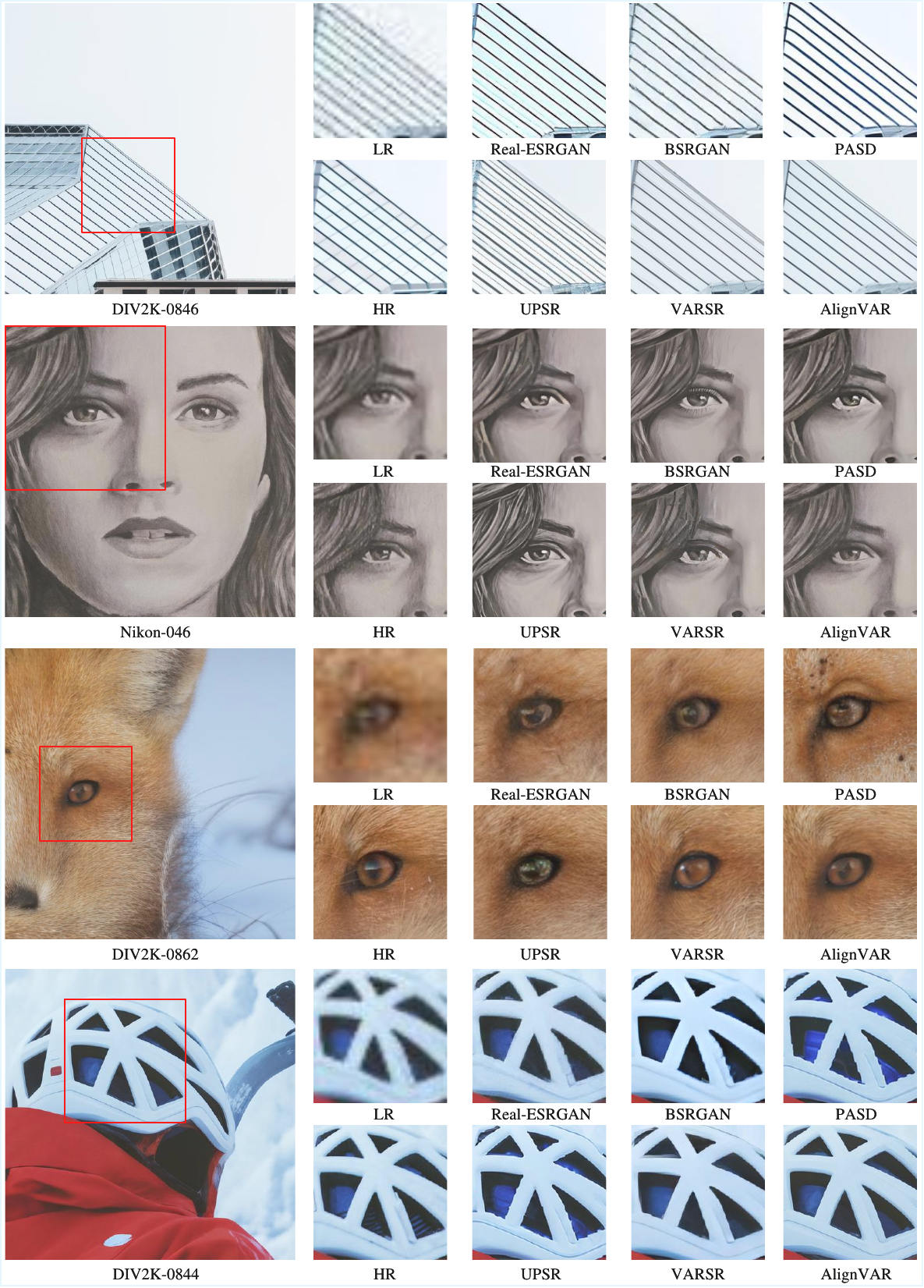}
    \caption{Additional qualitative comparisons. The large image on the left shows the output of AlignVAR, while the small patches on the right present enlarged crops from the LR input and competing methods.}
    \label{fig:visual_results2}
\end{figure*}
\vspace{-20pt}
\section{Additional Visual Results}
\label{sec:vis}
In this section, we present additional visual results to demonstrate the effectiveness of AlignVAR across various images. The results in Fig.~\ref{fig:visual_results1} and Fig.~\ref{fig:visual_results2} highlight the recovery of fine details in different scenes, with comparisons between AlignVAR and other state-of-the-art models, including GAN-based models (Real-ESRGAN~\cite{wang2021realesrgan}, BSRGAN~\cite{zhang2021bsrgan}) and diffusion-based models (PASD~\cite{yang2024pasd}, UPSR~\cite{zhang2025upsr}).

In Fig.~\ref{fig:visual_results1}, AlignVAR performs notably better in recovering intricate details such as the beak of the penguin, architectural structures. GAN-based models like BSRGAN, in contrast, tend to generate overly smooth results, losing fine details, especially in intricate structures like the penguin's beak or the architectural details. Diffusion models such as PASD also struggle with fine texture recovery, often resulting in overly blurred outputs, especially in high-frequency areas. However, compared to the HR image, some gaps remain, especially in areas with complex texture. This is likely due to the difficulty in recovering fine details, such as small textures in challenging low-resolution inputs. In Fig.~\ref{fig:visual_results2}, AlignVAR excels in recovering sketch-like images and animal details, such as the eye and fur texture of the animal. The eyes are restored with sharpness, natural shine, and detailed textures that are closer to the ground truth compared to other methods. GAN models like BSRGAN tend to oversmooth such textures, leading to unnatural results, while diffusion models again introduce unnecessary blurring, especially in small, detailed areas like the eyes or fur.

\section{Limitation}
\label{sec:limit}
While AlignVAR achieves strong performance across a wide range of benchmarks, several limitations remain. First, the recovery of middle scales proves to be particularly challenging, as shown in our hyperparameter analysis \ref{sec:hyperparameter}. These intermediate resolutions more sensitive to hyperparameter choices such as $\lambda$. In future work, we plan to explore dynamically adjustable consistency constraints that adapt to the difficulty of each scale. Second, although AlignVAR improves the spatial and hierarchical consistency of autoregressive prediction, its computation is still dominated by the final high-resolution scale, inheriting the $O(n^4)$ complexity of VAR-based models. This may limit scalability to extremely high-resolution image generation. Finally, our method relies on the quality of VQ tokenization. When the codebook fails to accurately represent rare textures or subtle structures, even consistent autoregression cannot fully recover the missing details. Incorporating stronger generative priors or adaptive codebook refinement may further enhance robustness to challenging degradations.


\end{document}